\documentclass{article} 
\usepackage{iclr2024_conference,times}

\usepackage{amsmath,amsfonts,bm}

\def\eqref#1{equation~\ref{#1}}

\def\1{\bm{1}}

\DeclareMathAlphabet{\mathsfit}{\encodingdefault}{\sfdefault}{m}{sl}
\SetMathAlphabet{\mathsfit}{bold}{\encodingdefault}{\sfdefault}{bx}{n}

\usepackage{hyperref}
\usepackage{url}

\usepackage{algorithm}
\usepackage{algorithmic}

\usepackage{multirow}
\usepackage{subfig}
\usepackage{float}
\usepackage{multicol}

\usepackage{amsmath}
\usepackage{amsthm}
\usepackage{booktabs}
\usepackage{algorithm}
\usepackage{algorithmic}
\usepackage[switch]{lineno}
\usepackage{multirow}
\usepackage{graphicx}
\usepackage{subfig}
\usepackage{xcolor}
\usepackage{amsfonts}
\usepackage[scientific-notation=true]{siunitx}

\usepackage{newfloat}
\usepackage{listings}

\title{Advancing Enterprise Spatio-Temporal Forecasting Applications: Data Mining Meets Instruction Tuning of Language Models for Multi-Modal Time Series Analysis in Low-Resource Settings}

\author{
Sagar Srinivas Sakhinana\thanks{Corresponding author} \\
\texttt{sagar.sakhinana@tcs.com} \\
\texttt{TCS  Research}
\And
Geethan Sannidhi \\
\texttt{geethan.iiitp.ac.in} \\
\texttt{IIIT Pune} 
\AND
Chidaksh Ravuru\\
\texttt{200010046@iitdh.ac.in} \\
\texttt{IIT Dharwad} 
\And
Venkataramana Runkana \\
\texttt{venkat.runkana@tcs.com} \\
\texttt{TCS  Research}
}

\iclrfinalcopy 
\begin{document}

\maketitle
\vspace{-6mm}
\begin{abstract}
\vspace{-4mm}
Spatio-temporal forecasting is crucial in transportation, logistics, and supply chain management. However, current methods struggle with large, complex datasets. We propose a dynamic, multi-modal approach that integrates the strengths of traditional forecasting methods and instruction tuning of small language models for time series trend analysis. This approach utilizes a mixture of experts (MoE) architecture with parameter-efficient fine-tuning (PEFT) methods, tailored for consumer hardware to scale up AI solutions in low resource settings while balancing performance and latency tradeoffs. Additionally, our approach leverages related past experiences for similar input time series to efficiently handle both intra-series and inter-series dependencies of non-stationary data with a time-then-space modeling approach, using grouped-query attention, while mitigating the limitations of traditional forecasting techniques in handling distributional shifts. Our approach models predictive uncertainty to improve decision-making. Our framework enables on-premises customization with reduced computational and memory demands, while maintaining inference speed and data privacy/security. Extensive experiments on various real-world datasets demonstrate that our framework provides robust and accurate forecasts, significantly outperforming existing methods.
\vspace{-7mm}
\end{abstract}

\vspace{-5mm}
\section{Introduction}
\begin{figure}[ht!]
\centering
\resizebox{0.9\linewidth}{!}{ 
\hspace*{-10mm}\includegraphics[keepaspectratio,height=4.5cm,trim=0.0cm 3.7cm 0cm 3.2cm,clip]{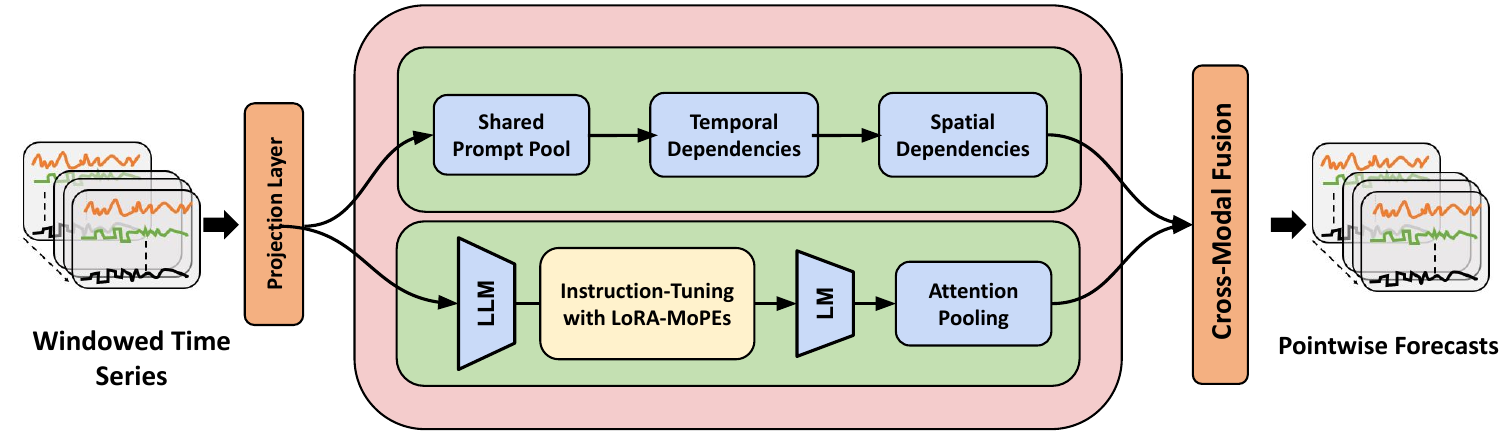} 
}
\vspace{-3mm}
\caption{The figure above illustrates our proposed comprehensive framework that synergizes three core strategies for advanced time series analysis. First, it employs a dynamic prompting mechanism that leverages historical learned patterns to adapt to emerging trends and capture dependencies both within and across time series to compute context-aware time series embeddings. Second, it fine-tunes a smaller language model using instruction-following data generated by larger models for time series trend analysis, yielding text-level embeddings that encapsulate these patterns. Lastly, it integrates these diverse, complementary cross-modal embeddings, offering accurate forecasts and improved generalization and scalability for practical forecasting applications. The model architecture is explained in great detail in Section\ref{method}. Refer to the technical appendix for more information.}
\label{fig:OverallArch}
\vspace{-5mm}
\end{figure}

\vspace{-4mm}
Multivariate time series forecasting (MTSF) has many applications, but it faces challenges such as complex relationships between time series variables, non-linearity, sparsity, and non-stationarity. Spatio-temporal graph neural networks (STGNNs) improve forecast accuracy by modeling temporal dependencies within variables and interdependencies between variables. STGNNs utilize both explicit relationships based on predefined graphs provided by domain expert knowledge and implicit relationships derived from data-driven relational inference methods. While `Human-in-the-loop' STGNNs \cite{yu2017spatio, li2017diffusion, guo2020optimized} use prior knowledge in predefined graphs, they do not take into account latent variable relationships underlying the substantial data. On the other hand, `Human-out-of-the-loop' STGNNs \cite{deng2021graph, wu2020connecting, kipf2018neural} jointly infer variable dependency graph structures and learn spatio-temporal dynamics from data, but they may underutilize expert-defined graphs, especially in noisy data scenarios, which can impact forecasting performance. However, existing methods rely on fixed historical window lengths that may not capture the diverse and complex time series patterns of varying lengths, and lack reliable uncertainty estimates. Transformers\cite{vaswani2017attention}, without a built-in bias towards pairwise variable dependencies, provide greater flexibility in modeling long-range dependencies beyond local spatial relationships, enabling the capture of global relationships. STGNNs introduce a stronger spatio-temporal inductive bias, while Transformers provide enhanced representational flexibility. Recent research indicates an opportunity to develop hybrid methods that combine explicit domain knowledge in priori-known graph structures with data-driven relational learning, overcome the limitations of fixed-length window sequences by enabling the forecasting methods to apply learned patterns across past observations, and provide probabilistic forecasts for more accurate and reliable MTSF. In recent years, proprietary and closed-source large language models (LLMs) such as GPT-4 \cite{gpt4} have revolutionized natural language processing by achieving remarkable performance through pretraining on diverse and massive datasets. However, their black-box nature hinders interpretability in applications. Open-source LLMs, such as Llama 2 \cite{touvron2023llama}, allow fine-tuning for domain-specific customization but require significant computational resources. In contrast, smaller open-source models, like BERT \cite{devlin2018bert}, are interpretable but may lack reasoning abilities compared to contemporary off-the-shelf LLMs. Furthermore, integrating foundational LLMs with traditional forecasting methods remains largely unexplored, yet it holds great promise for enhancing predictions. Adapting LLMs to generate natural language descriptions capturing time series trends and patterns, though unconventional, offers a clear possibility for providing unique insights that complement and guide traditional forecasting techniques. However, sharing sensitive data with external proprietary LLM API services raises concerns regarding data privacy, sovereignty, costs, and security and has limited ability to customize them for specific needs. In this work, we introduce \texttt{MultiTs} Net, an innovative dynamic, multi-modal approach that combines prompt-based time series representation learning with complementary instruction-tuning open-source language models for time series trend analysis, aiming to enhance accurate time series forecasting. For an illustration of the proposed framework, please refer to Figure \ref{fig:OverallArch}. The key design methods include: (a) utilizing a flexible retrieval-based prompt pool to integrate time series-specific knowledge and historical context relevant to the current data distribution into traditional forecasting methods. (b) Instruction tuning of small-scale language models for time series trend analysis to interpret and describe complex time series data. The prompt design involves creating a shared pool of prompts stored as distinct key-value pairs. These prompts are tailored to encode task-specific knowledge, such as trends or seasonality relevant to different time periods. The framework uses these prompts to recognize and apply learned patterns to guide its predictions for each time series instance through transfer learning. Our framework leverages related past experiences, where similar input time series instances retrieve the same group of prompts from the pool, enhancing efficiency and predictive performance. This approach is particularly suited for overcoming fixed-window size limitations by handling the typically nonstationary nature of real-world time series data with distributional shifts. The traditional representation learning method captures the full spectrum of both intra- and inter-series dependencies underlying the time series data by implementing a time-then-space modeling approach\cite{gao2022equivalence}, focusing on temporal dynamics before learning spatial dependencies. We utilize a grouped-query attention mechanism\cite{ainslie2023gqa} to learn long-range temporal dependencies. The spatial learning method includes: (i) Graph Chebyshev convolutions to leverage explicit prior knowledge of domain expert-based pairwise variable dependencies. (ii) Utilizing grouped-query attention with no graph spatial priors to learn all pairwise variable dependencies, enhancing the understanding of long-range spatial dependencies. We perform a convex combination through a gating mechanism to compute accurate latent representations of the complex non-linear spatial dynamics of the time series data, which improves forecasting accuracy. We utilize a large-scale open-source LLM, such as `llama2 70B 4k,' to generate instruction-following data consisting of pairs of time-series data and the corresponding natural language descriptions that encapsulate insights into time-series trends and patterns. We employ the Mixture of Parameter-Efficient Experts (MoPEs) with Low-Rank Adaptation (LoRA) technique for on-premises instruction-tuning of small-scale decoder-only language models, such as `llama2 7B 4k,' using this machine-generated data to customize them for time-series trend analysis for enterprise adoption to complement traditional forecasting methods. In addition, the framework models predictive uncertainty to assist decision-making. In summary, our proposed framework integrates open-source large and small-scale language models with traditional time series representation learning methods by dynamically adapting to the evolving nature of non-stationary time series data distributions to provide robust and accurate forecasts. It offers a secure and affordable solution to run on consumer-grade hardware within their infrastructure, enhancing data privacy and reducing costs, thereby enabling the realization of the benefits of language models for time series analysis while addressing key adoption barriers. For more information on the proposed framework, refer to Section\ref{method} in the technical appendix.

\vspace{-6mm}
\section{Problem Definition}
\label{definition}
\vspace{-4mm}
Our study focuses on a dynamic system with \( N \) sensors collecting sequential data over \( T \) time intervals across \( F \) features, forming a spatio-temporal matrix \( \mathbf{X} \in \mathbb{R}^{N \times T \times F} \). These features include traffic attributes, such as speed, flow, and density. We denote the historical data for each sensor as \( \mathbf{X}_{i} \in \mathbb{R}^{T \times F} \), and the data for all sensors at each time step as \( \mathbf{X}^{t} \in \mathbb{R}^{N \times F} \). To improve our framework's ability to adapt to traffic patterns, we apply a sliding window technique to segment the historical data into overlapping, consecutive samples \( \mathbf{X}^{t-W : t-1} \in \mathbb{R}^{N \times W \times F} \), where \( W \) is the window size over past observations. We focus on a single feature, specifically, \( F=1 \),  to accurately predict future traffic flow, allowing for consistent comparison with baseline models. Our neural network model, denoted as \( \boldsymbol{\Theta} \), aims to forecast traffic data for the upcoming \(\nu\) steps into the future, represented as \( \mathbf{S}^{t+1} = \mathbf{X}^{t: t + \nu - 1} \), based on past observations, which are denoted as \( \mathbf{S}^{t} = \mathbf{X}^{t-W: t-1} \).
We train the framework using a mean absolute error (MAE) loss function as follows, 

\vspace{0mm}
\resizebox{0.925\linewidth}{!}{
\hspace{0mm}\begin{minipage}{\linewidth}
\begin{equation}
\mathcal{L}(\mathbf{\Theta}) = \frac{1}{|\nu|} |\hat{\mathbf{S}}^{t+1} - \mathbf{S}^{t+1}| \nonumber
\end{equation}
\end{minipage}
}

where $\hat{\mathbf{S}}^{t+1}$ denote the framework predictions.

\vspace{-4mm} 

\vspace{-2mm}
\section{Experiments and Results}
\label{results}

\vspace{-5mm}
\begin{table}[tbhp]
\center
\setlength{\tabcolsep}{0.25em} 
\renewcommand\arraystretch{1.325} 
\centering
\resizebox{0.6\textwidth}{!}{
\hspace{-5mm}\begin{tabular}{c|c|c|c|c|c}
\hline
\textbf{Dataset} & \textbf{Nodes} & \textbf{Timesteps} & \textbf{Time-Range} & \multicolumn{1}{l|}{\textbf{Data Split}} & \multicolumn{1}{l}{\textbf{Granularity}} \\ \hline
PeMSD3 & 358 & 26,208 & 09/2018 - 11/2018 & \multirow{5}{*}{6 / 2 / 2} & \multirow{7}{*}{\rotatebox[origin=c]{270}{5 mins}} \\
PeMSD4 & 307 & 16,992 & 01/2018 - 02/2018 &  &  \\
PeMSD7 & 883 & 28,224 & 05/2017 - 08/2017 &  &  \\
PeMSD8 & 170 & 17,856 & 07/2016 - 08/2016 &  &  \\
PeMSD7(M) & 228 & 12,672 & 05/2012 - 06/2012 &  &  \\ \cline{1-5}
METR-LA & 207 & 34,272 & 03/2012 - 06/2012 & \multirow{2}{*}{7 / 1 / 2} &  \\
PEMS-BAY & 325 & 52,116 & 01/2017 - 05/2017 &  &  \\ \hline
\end{tabular}
}
\vspace{-2mm}
\caption{The table provides a comprehensive overview of the benchmark traffic datasets used for the MTSF task, highlighting the timesteps, time range, train/validation/test split.}
\label{tab:summarydatasets}
\vspace{-4mm}
\end{table} 

\vspace{-4mm}
\subsection{Datasets}
\vspace{-3mm}
The study focuses on evaluating two frameworks: \texttt{MultiTs} Net and its variant \texttt{w/Unc-MultiTs} Net, using large-scale spatial-temporal traffic datasets (PeMSD3, PeMSD4, PeMSD7, PeMSD7(M), PeMSD8) from the Caltrans Performance Measurement System (PeMS)\cite{chen2001freeway}. PeMS provides critical real-time and historical traffic data for California's freeways, aiding in traffic management, monitoring, and analysis. Our study converts 30-second interval data into 5-minute averages, following previous research methods\cite{choi2022graph}, and also uses additional traffic flow datasets (METR-LA and PEMS-BAY)\cite{LiYS018} converted into the same format. This approach allows for a consistent data format, enhancing the study's ability to analyze and model complex spatial-temporal data, and demonstrate superior performance over existing methodologies. Table \ref{tab:summarydatasets} summarizes details about the benchmark datasets used in this study.

\vspace{-4.5mm}
\subsection{Experimental Setup}
\vspace{-3mm}
In our study, we divided traffic-related datasets (PEMS-BAY and METR-LA) into three parts: training (70$\%$), validation (10$\%$), and testing (20$\%$). Other datasets followed a 60\%/20\%/20\% split. Before training, we standardized all time-series variables to a zero mean and unit variance. The models were assessed using Mean Absolute Error (MAE), Root Mean Squared Error (RMSE), and Mean Absolute Percentage Error (MAPE), calculated on the original scale of data. Key hyperparameters included: window size ($W$ = 12), prompt pool size ($M$ = 15), and embedding dimension ($d$ = 64). The Grouped-query multi-head attention (GQ-MHA) parameters included the number of groups ($g$ = 3), attention heads ($H$ = 4), and key/value/query dimensions ($d_{h}$ = 16), balancing model accuracy and efficiency. Our \texttt{MultiTs} Net framework was trained over 30 epochs with a batch size of 48, employing early stopping based on Validation MAE to prevent overfitting. A learning rate scheduler and the Adam optimizer were used for efficient training, with the initial rate set at 1e-3. Training involved powerful NVIDIA Tesla V100 GPUs and multiple experimental runs, reporting ensemble averages for model performance evaluation. We minimized MAE loss for \texttt{MultiTs} Net and Gaussian negative log-likelihood loss for its variant \texttt{w/Unc-MultiTs} Net. For instruction-tuning the Llama2-7B model involved utilizing the MoPEs technique with key hyperparameters: rank ($r$ = 16), controlling model capacity; alpha ($\alpha$ = $\frac{1}{16}$, a fraction of the rank), determining parameter update magnitude; and a LoRA dropout rate (0.05) for generalization. Training settings included a batch size of 16 per GPU, 15 epochs, an initial learning rate of 2e-4, weight decay of 0.001, the AdamW optimizer, and 8-bit quantization via MoPEs for efficiency. The focus was on supervised training to minimize the cross-entropy loss aiming to minimize cross-entropy loss by correlating time series data with textual descriptions using Nvidia V100 GPUs and the PyTorch library.

\vspace{-7mm}
\hspace{-10mm} \begin{table*}[ht!]
\centering
\setlength{\tabcolsep}{0.21em} 
\renewcommand\arraystretch{1.075} 
 \resizebox{1.05\textwidth}{!}{
\hspace{-10mm} \begin{tabular}{c|ccc|ccc|ccc|ccc|ccc}
\hline
\multirow{2}{*}{\textbf{Methods}} & \multicolumn{3}{c|}{\textbf{PeMSD3}} & \multicolumn{3}{c|}{\textbf{PeMSD4}} & \multicolumn{3}{c|}{\textbf{PeMSD7}} & \multicolumn{3}{c|}{\textbf{PeMSD8}} & \multicolumn{3}{c}{\textbf{PeMSD7(M)}} \\ \cline{2-16} 
 & \multicolumn{1}{l}{\textbf{MAE}} & \multicolumn{1}{l}{\textbf{RMSE}} & \multicolumn{1}{l|}{\textbf{MAPE}} & \multicolumn{1}{l}{\textbf{MAE}} & \multicolumn{1}{l}{\textbf{RMSE}} & \multicolumn{1}{l|}{\textbf{MAPE}} & \multicolumn{1}{l}{\textbf{MAE}} & \multicolumn{1}{l}{\textbf{RMSE}} & \multicolumn{1}{l|}{\textbf{MAPE}} & \multicolumn{1}{l}{\textbf{MAE}} & \multicolumn{1}{l}{\textbf{RMSE}} & \multicolumn{1}{l|}{\textbf{MAPE}} & \textbf{MAE} & \textbf{RMSE} & \textbf{MAPE} \\ \hline
HA & 31.58 & 52.39 & 33.78 & 38.03 & 59.24 & 27.88 & 45.12 & 65.64 & 24.51 & 34.86 & 59.24 & 27.88 & 4.59 & 8.63 & 14.35 \\
ARIMA & 35.41 & 47.59 & 33.78 & 33.73 & 48.80 & 24.18 & 38.17 & 59.27 & 19.46 & 31.09 & 44.32 & 22.73 & 7.27 & 13.20 & 15.38 \\
VAR & 23.65 & 38.26 & 24.51 & 24.54 & 38.61 & 17.24 & 50.22 & 75.63 & 32.22 & 19.19 & 29.81 & 13.10 & 4.25 & 7.61 & 10.28 \\
FC-LSTM & 21.33 & 35.11 & 23.33 & 26.77 & 40.65 & 18.23 & 29.98 & 45.94 & 13.20 & 23.09 & 35.17 & 14.99 & 4.16 & 7.51 & 10.10 \\
TCN & 19.32 & 33.55 & 19.93 & 23.22 & 37.26 & 15.59 & 32.72 & 42.23 & 14.26 & 22.72 & 35.79 & 14.03 & 4.36 & 7.20 & 9.71 \\
TCN(w/o causal) & 18.87 & 32.24 & 18.63 & 22.81 & 36.87 & 14.31 & 30.53 & 41.02 & 13.88 & 21.42 & 34.03 & 13.09 & 4.43 & 7.53 & 9.44 \\
GRU-ED & 19.12 & 32.85 & 19.31 & 23.68 & 39.27 & 16.44 & 27.66 & 43.49 & 12.20 & 22.00 & 36.22 & 13.33 & 4.78 & 9.05 & 12.66 \\
DSANet & 21.29 & 34.55 & 23.21 & 22.79 & 35.77 & 16.03 & 31.36 & 49.11 & 14.43 & 17.14 & 26.96 & 11.32 & 3.52 & 6.98 & 8.78 \\
STGCN & 17.55 & 30.42 & 17.34 & 21.16 & 34.89 & 13.83 & 25.33 & 39.34 & 11.21 & 17.50 & 27.09 & 11.29 & 3.86 & 6.79 & 10.06 \\
DCRNN & 17.99 & 30.31 & 18.34 & 21.22 & 33.44 & 14.17 & 25.22 & 38.61 & 11.82 & 16.82 & 26.36 & 10.92 & 3.83 & 7.18 & 9.81 \\
GraphWaveNet & 19.12 & 32.77 & 18.89 & 24.89 & 39.66 & 17.29 & 26.39 & 41.50 & 11.97 & 18.28 & 30.05 & 12.15 & 3.19 & 6.24 & 8.02 \\
ASTGCN(r) & 17.34 & 29.56 & 17.21 & 22.93 & 35.22 & 16.56 & 24.01 & 37.87 & 10.73 & 18.25 & 28.06 & 11.64 & 3.14 & 6.18 & 8.12 \\
MSTGCN & 19.54 & 31.93 & 23.86 & 23.96 & 37.21 & 14.33 & 29.00 & 43.73 & 14.30 & 19.00 & 29.15 & 12.38 & 3.54 & 6.14 & 9.00 \\
STG2Seq & 19.03 & 29.83 & 21.55 & 25.20 & 38.48 & 18.77 & 32.77 & 47.16 & 20.16 & 20.17 & 30.71 & 17.32 & 3.48 & 6.51 & 8.95 \\
LSGCN & 17.94 & 29.85 & 16.98 & 21.53 & 33.86 & 13.18 & 27.31 & 41.46 & 11.98 & 17.73 & 26.76 & 11.20 & 3.05 & 5.98 & 7.62 \\
STSGCN & 17.48 & 29.21 & 16.78 & 21.19 & 33.65 & 13.90 & 24.26 & 39.03 & 10.21 & 17.13 & 26.80 & 10.96 & 3.01 & 5.93 & 7.55 \\
AGCRN & 15.98 & 28.25 & 15.23 & 19.83 & 32.26 & 12.97 & 22.37 & 36.55 & 9.12 & 15.95 & 25.22 & 10.09 & 2.79 & 5.54 & 7.02 \\
STFGNN & 16.77 & 28.34 & 16.30 & 20.48 & 32.51 & 16.77 & 23.46 & 36.60 & 9.21 & 16.94 & 26.25 & 10.60 & 2.90 & 5.79 & 7.23 \\
STGODE & 16.50 & 27.84 & 16.69 & 20.84 & 32.82 & 13.77 & 22.59 & 37.54 & 10.14 & 16.81 & 25.97 & 10.62 & 2.97 & 5.66 & 7.36 \\
Z-GCNETs & 16.64 & 28.15 & 16.39 & 19.50 & 31.61 & 12.78 & 21.77 & 35.17 & 9.25 & 15.76 & 25.11 & 10.01 & 2.75 & 5.62 & 6.89 \\
STG-NCDE & 15.57 & 27.09 & 15.06 & 19.21 & 31.09 & 12.76 & 20.53 & 33.84 & 8.80 & 15.45 & 24.81 & 9.92 & 2.68 & 5.39 & 6.76 \\ \hline
\textbf{MultiTs Net} & \textbf{11.23} & \textbf{16.93} & \textbf{8.86} & \textbf{14.71} & \textbf{21.11} & \textbf{8.29} & \textbf{17.24} & \textbf{25.53} & \textbf{7.43} & \textbf{12.31} & \textbf{17.91} & \textbf{6.38} & \textbf{2.08} & \textbf{4.13} & \textbf{4.93} \\
\textbf{W/Unc-MultiTs Net} & \textbf{11.44} & \textbf{16.92} & \textbf{9.15} & \textbf{15.26} & \textbf{23.16} & \textbf{8.43} & \textbf{16.88} & \textbf{26.75} & \textbf{8.56} & \textbf{12.42} & \textbf{19.11} & \textbf{6.64} & \textbf{2.08} & \textbf{4.10} & \textbf{5.24} \\ \hline
\end{tabular}
}
\vspace{-2.5mm}
\caption{The table shows the multi-metric error estimations for a 12-sequence-to-12-sequence forecasting task on the benchmark datasets, PeMSD3, PeMSD4, PeMSD7, PeMSD8, and PeMSD7(M). A lower error indicates better model performance.}
\label{tab:results1}
\vspace{-5mm}
\end{table*}

\vspace{-5mm}
\subsection{Multistep Forecasting Results} 
\begin{figure*}[ht!]
\centering
\hspace*{0cm}\resizebox{0.875\textwidth}{!}{
\subfloat[MAPE on PeMSD7]{\includegraphics[width=50mm]{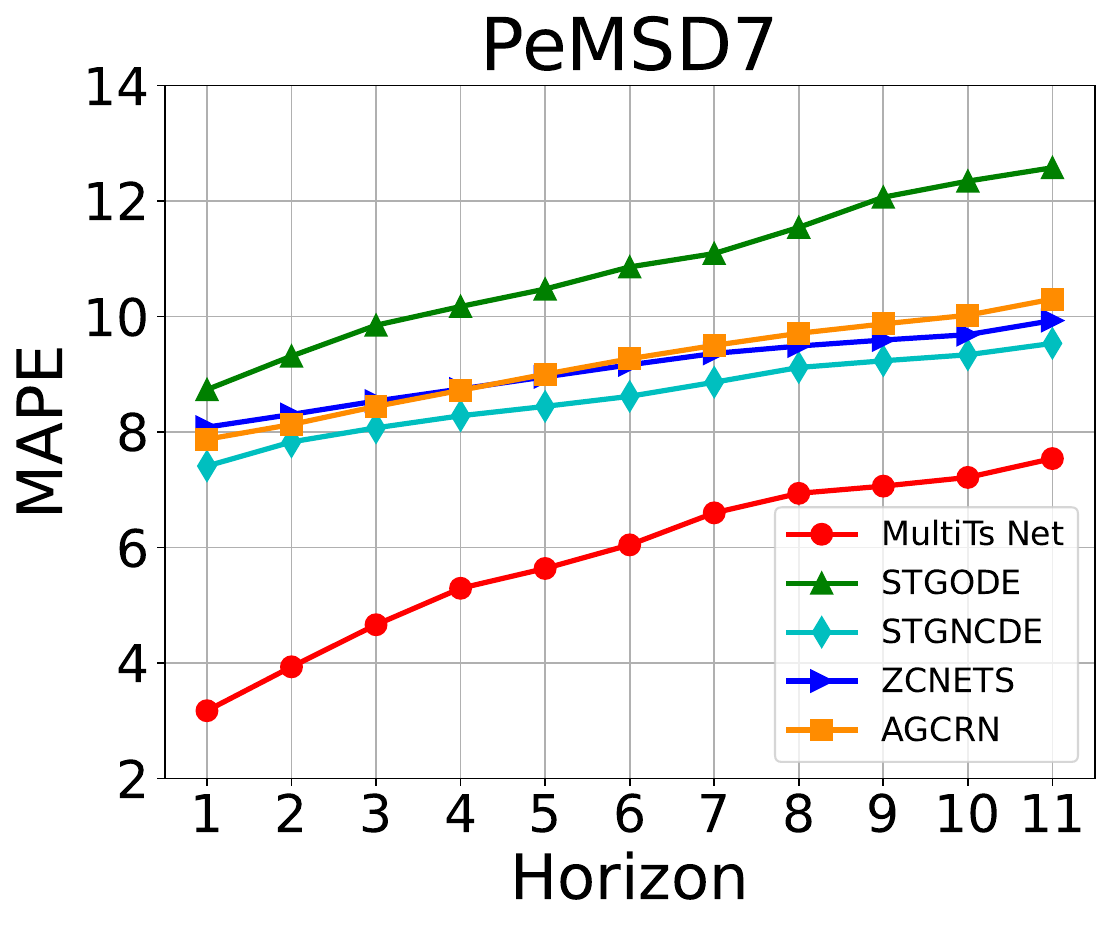}}
\subfloat[RMSE on PeMSD7]{\includegraphics[width=50mm]{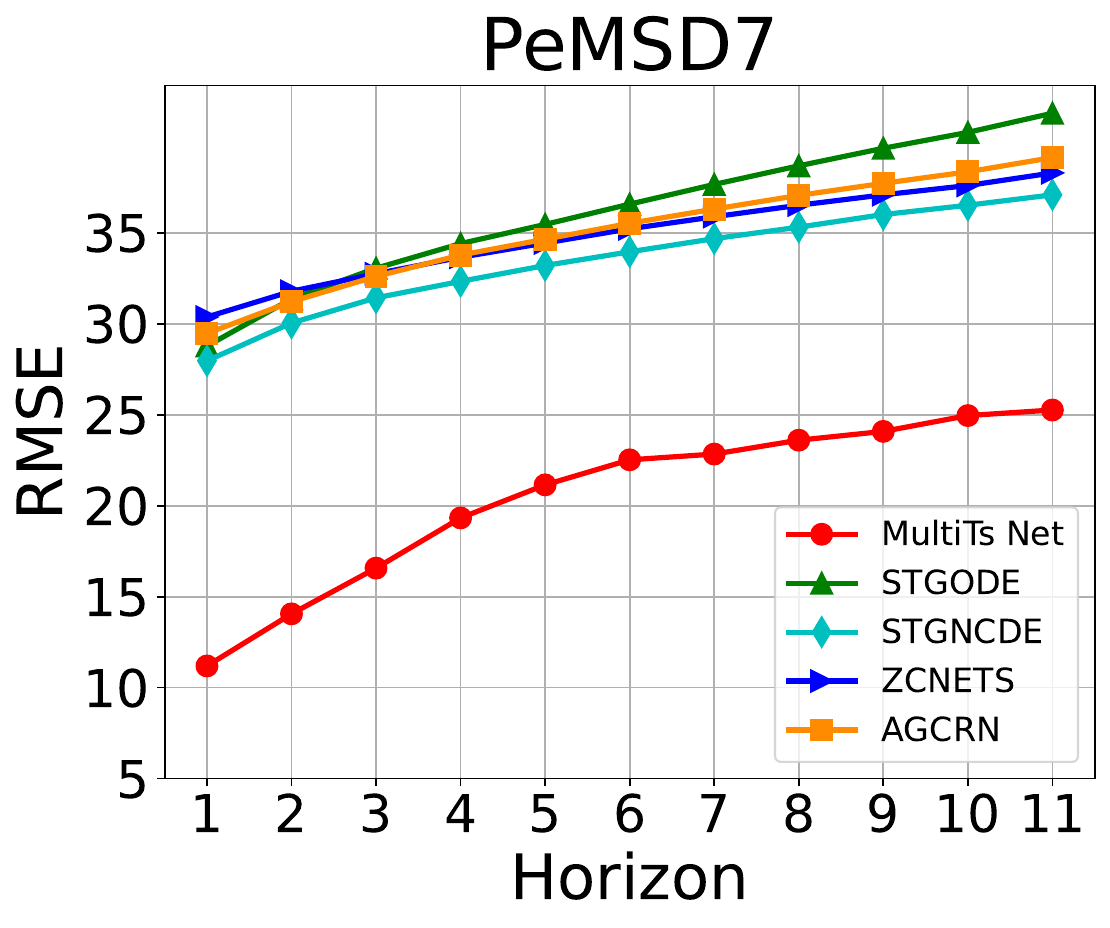}}
\subfloat[MAE on PeMSD7]{\includegraphics[width=50mm]{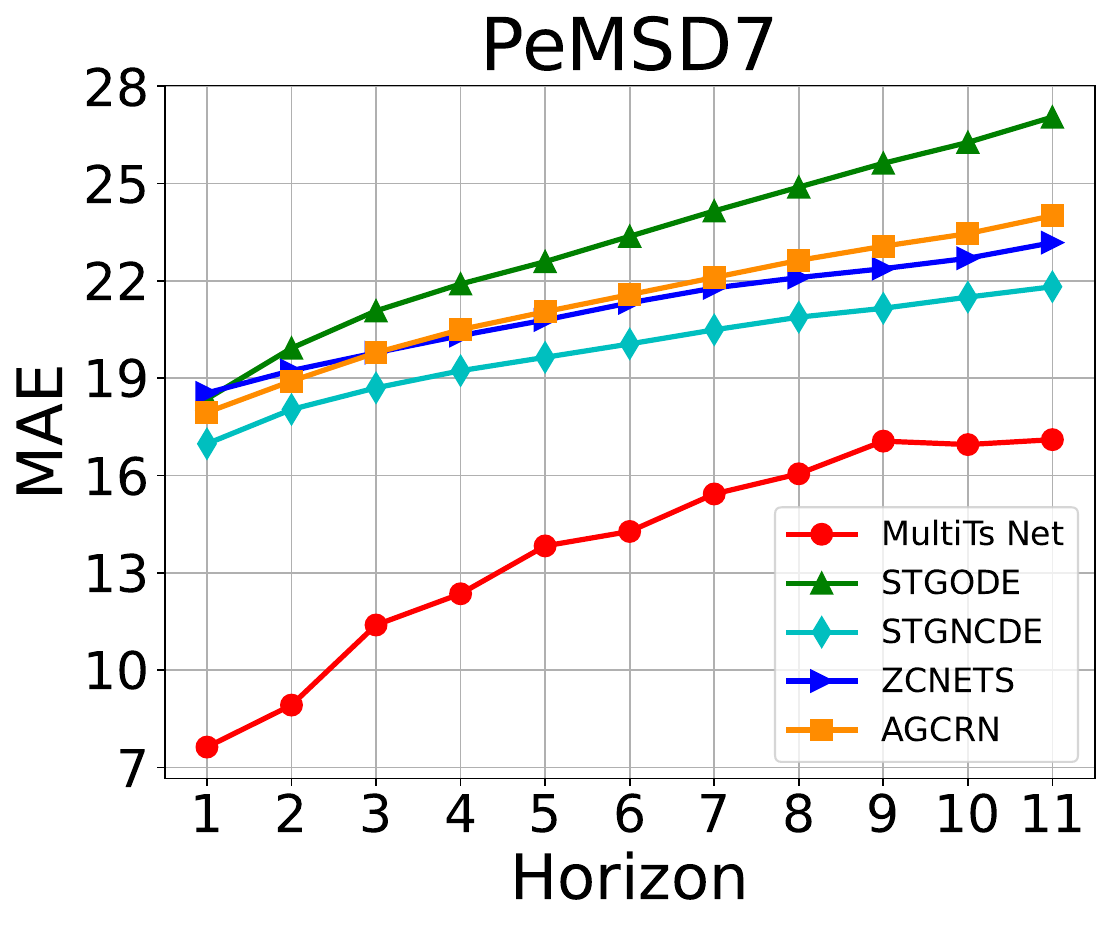}}
}\\[-2ex]
\hspace*{0cm}\resizebox{0.875\textwidth}{!}{
\subfloat[MAE on PeMSD4]{\includegraphics[width=50mm]{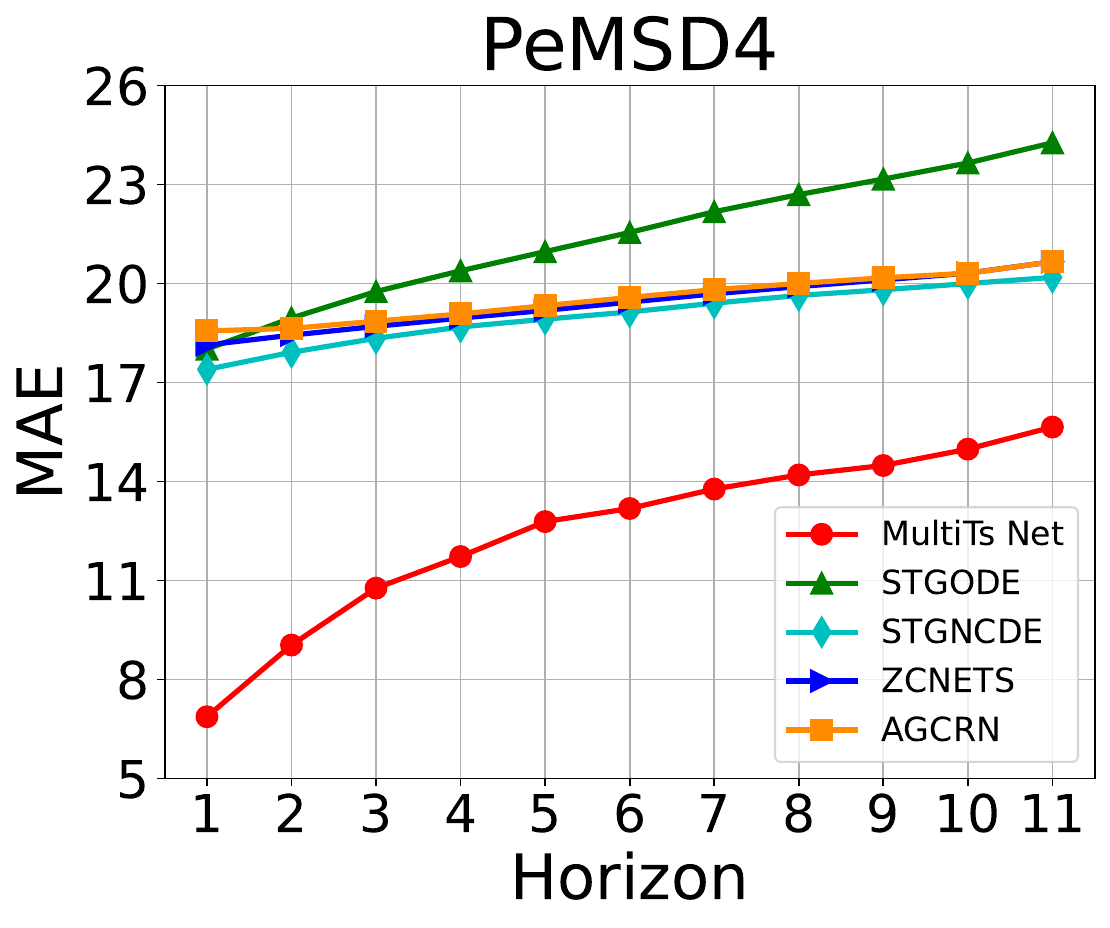}}
\subfloat[MAPE on PeMSD4]{\includegraphics[width=50mm]{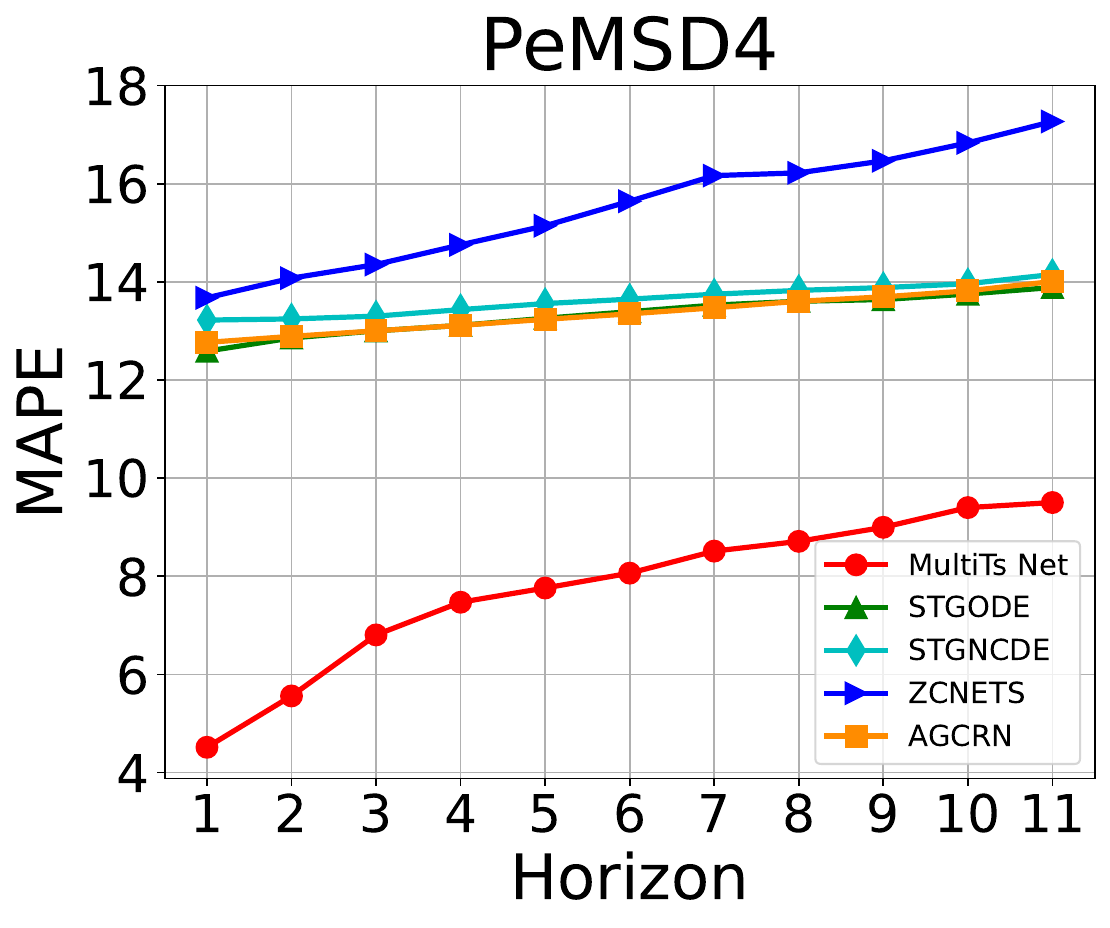}}
\subfloat[RMSE on PeMSD4]{\includegraphics[width=50mm]{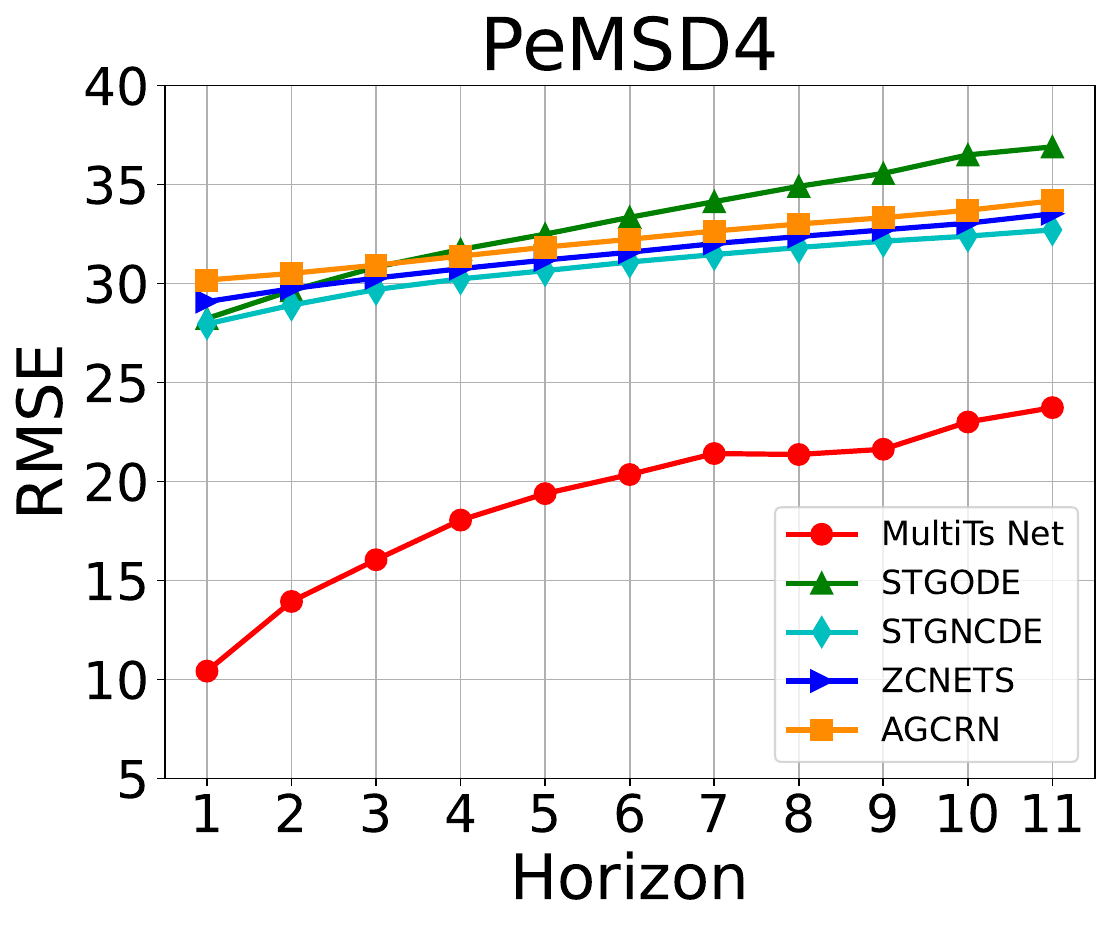}}
}\\[-2ex]
\hspace*{0cm}\resizebox{0.875\textwidth}{!}{
\subfloat[RMSE on PeMSD7(M)]{\includegraphics[width=50mm]{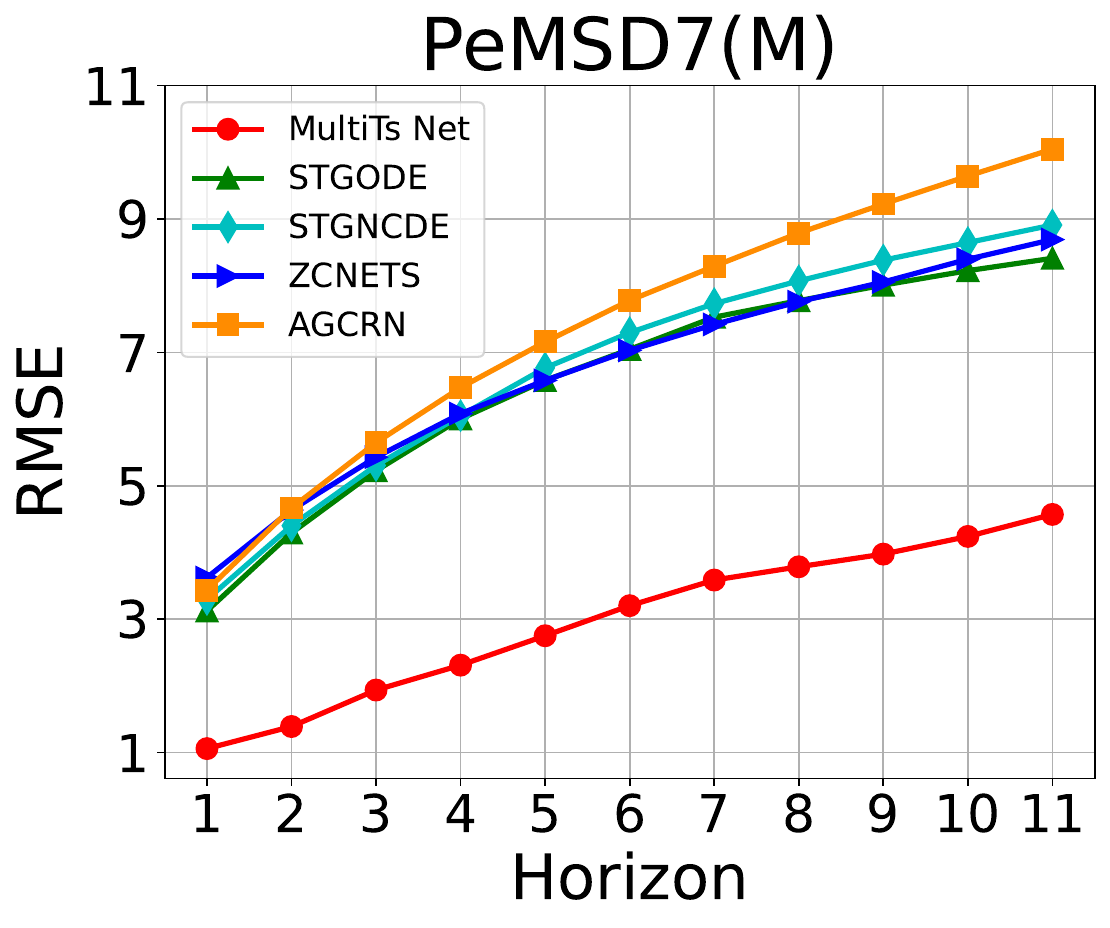}}
\subfloat[MAE on PeMSD7(M)]{\includegraphics[width=50mm]{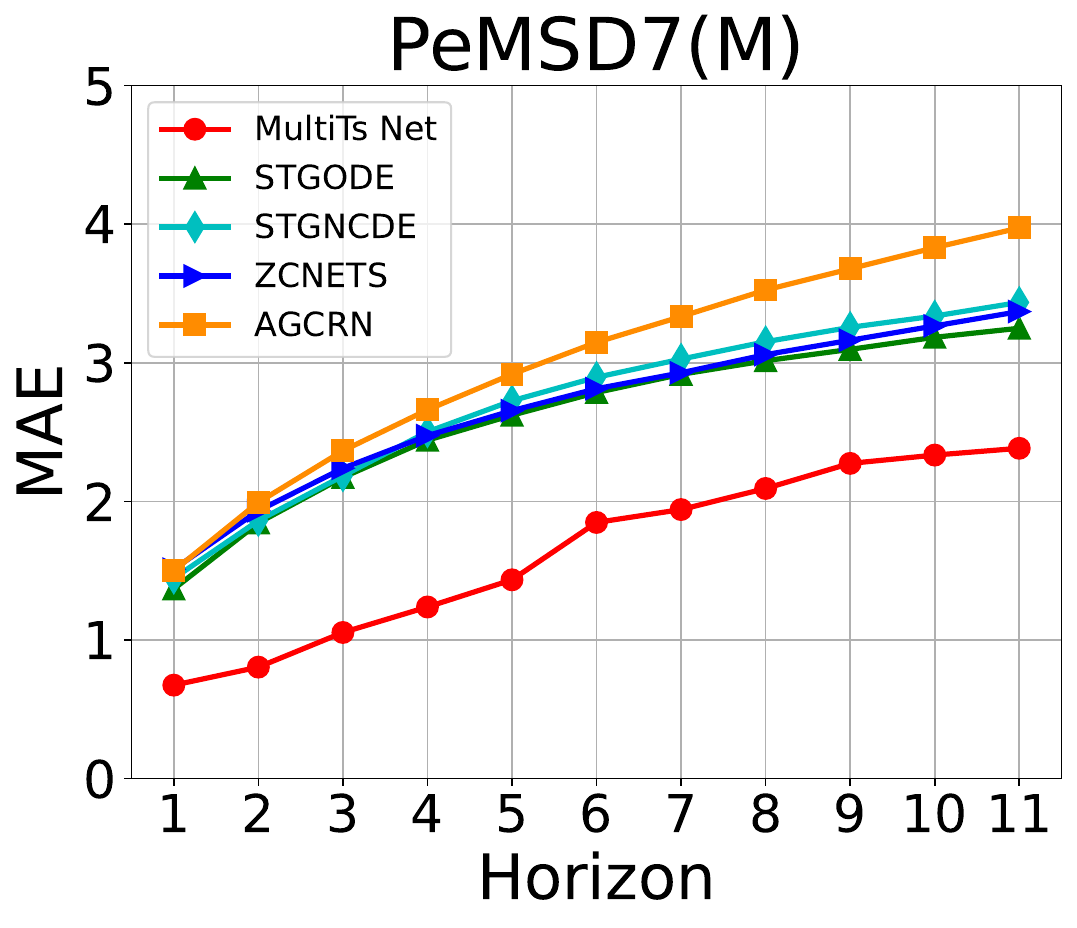}}
\subfloat[MAPE on PeMSD7(M)]{\includegraphics[width=50mm]{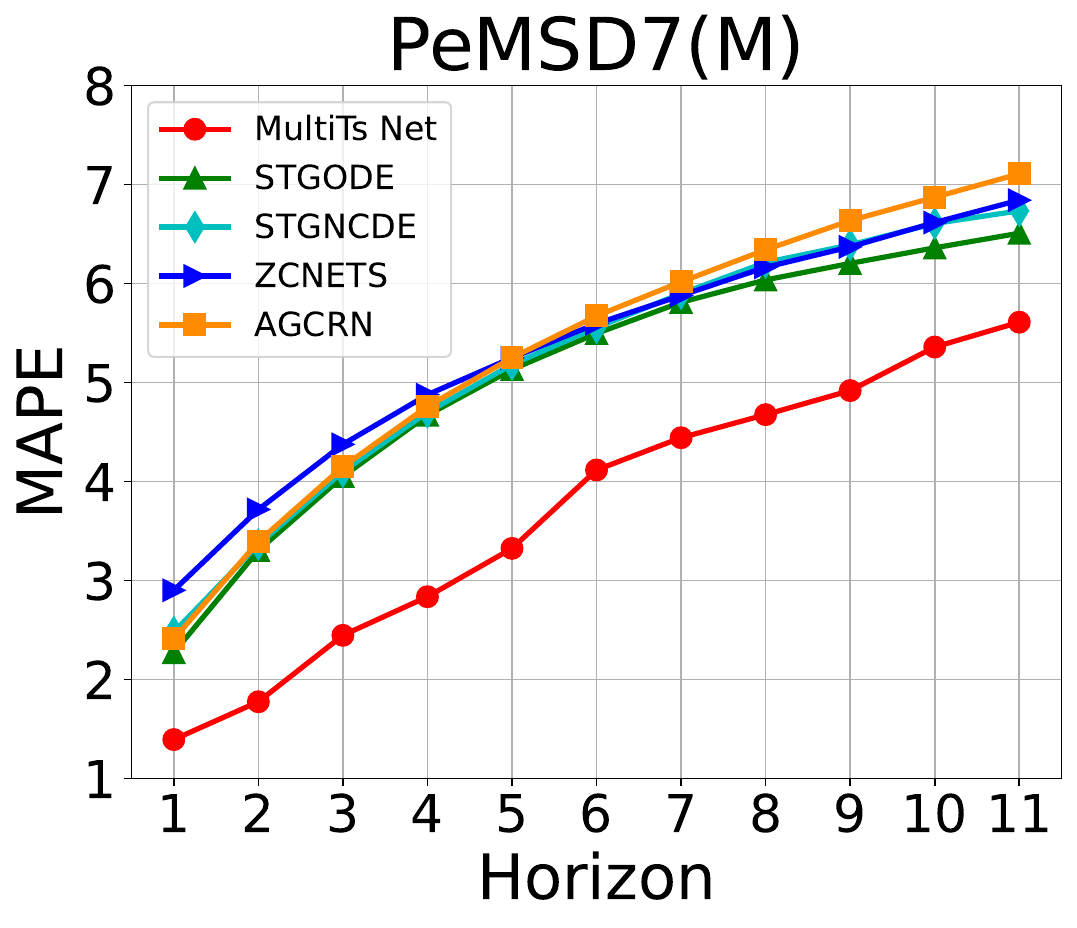}}
}\\[-2ex]
\hspace*{0cm}\resizebox{0.875\textwidth}{!}{
\subfloat[MAE on PeMSD3]{\includegraphics[width=50mm]{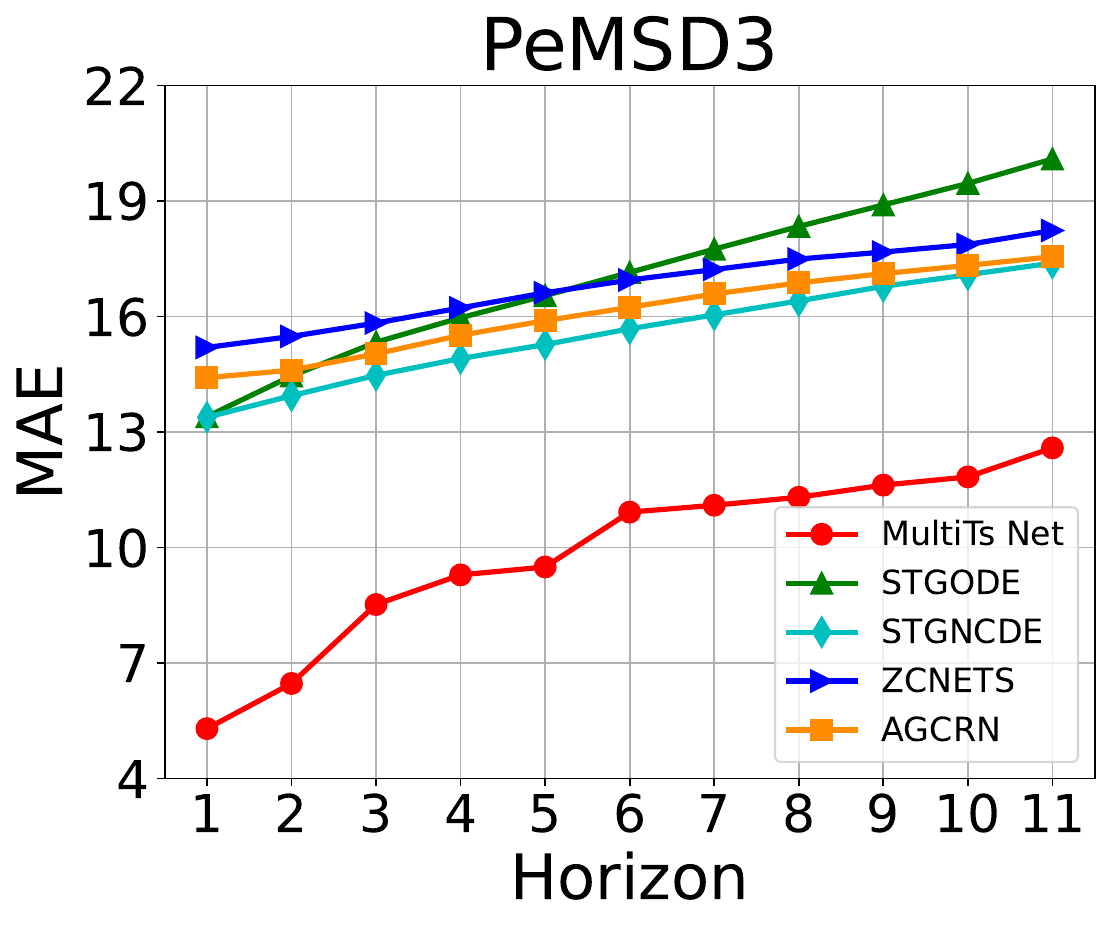}}
\subfloat[MAPE on PeMSD3]{\includegraphics[width=50mm]{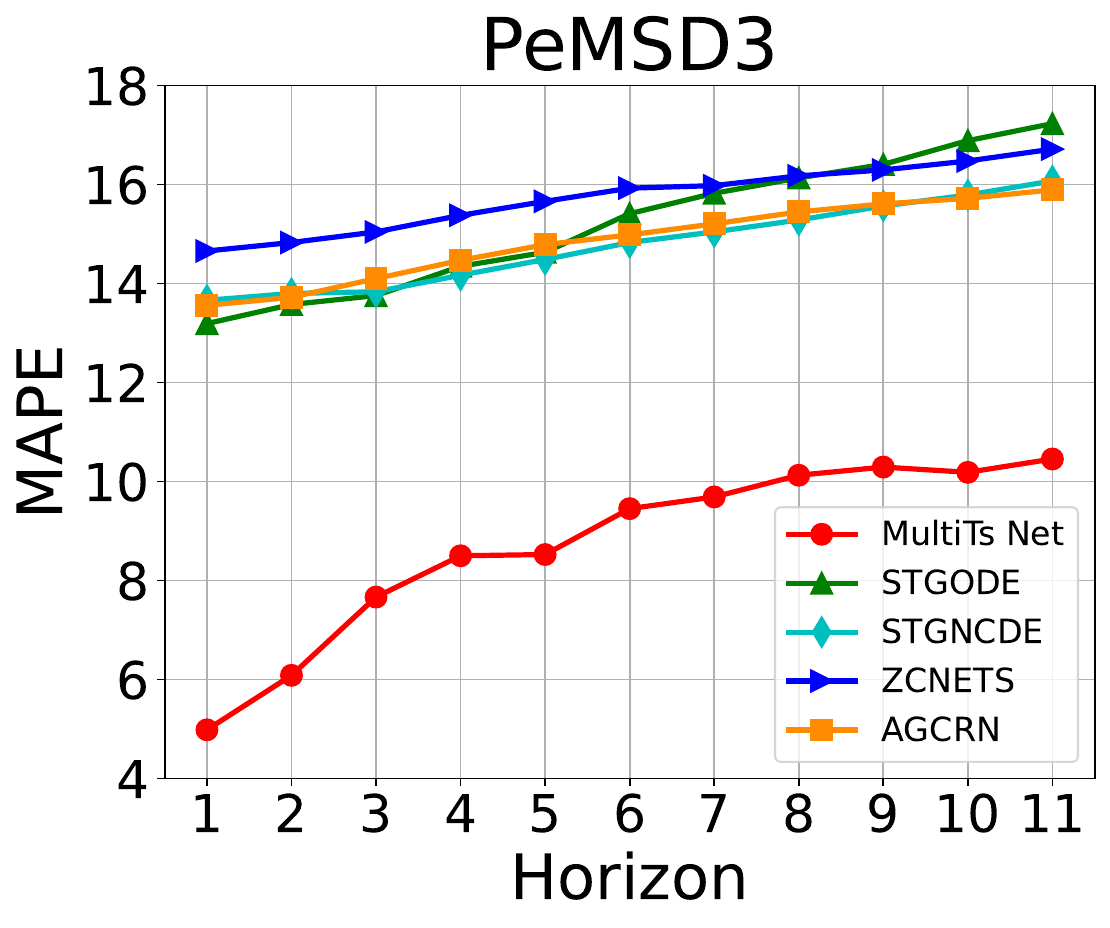}}
\subfloat[RMSE on PeMSD3]{\includegraphics[width=50mm]{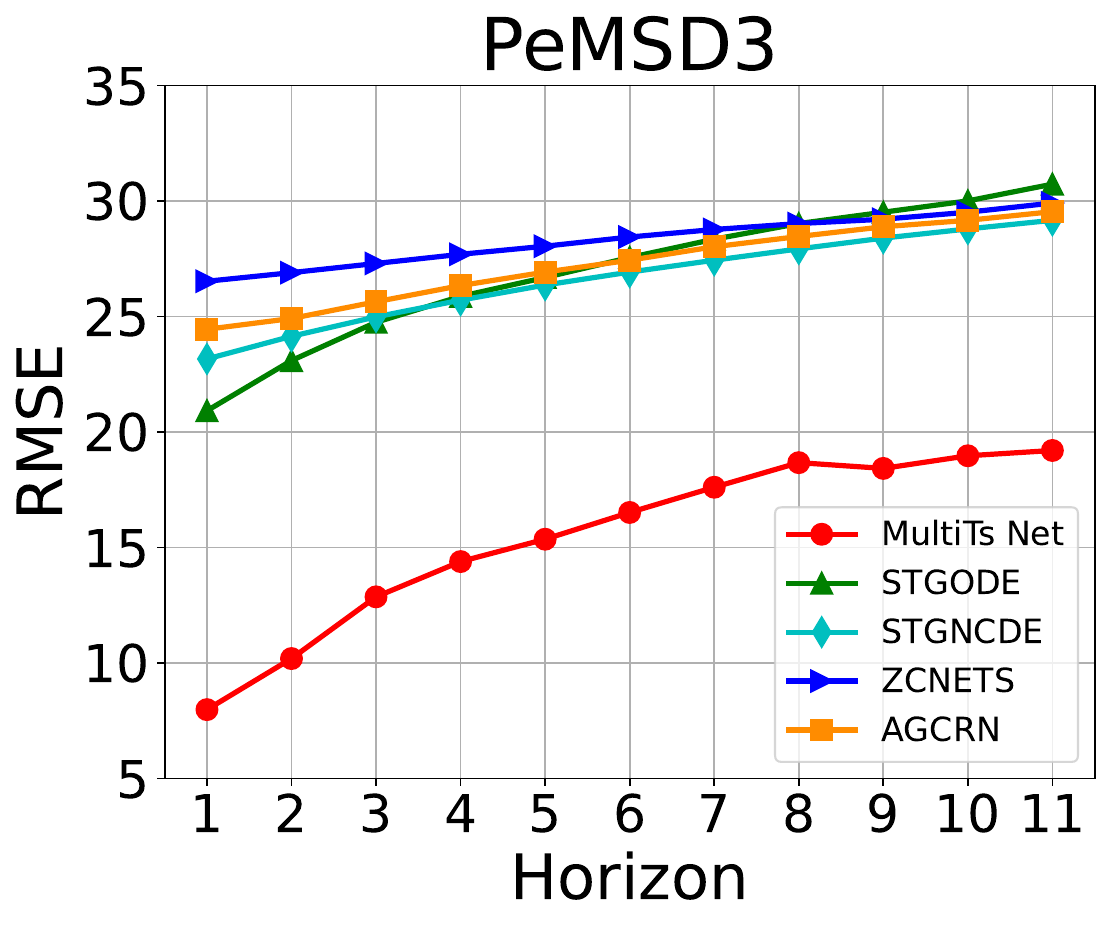}}
}\\[-2ex]
\hspace*{0cm}\resizebox{0.875\textwidth}{!}{
\subfloat[MAE on PeMSD8]{\includegraphics[width=50mm]{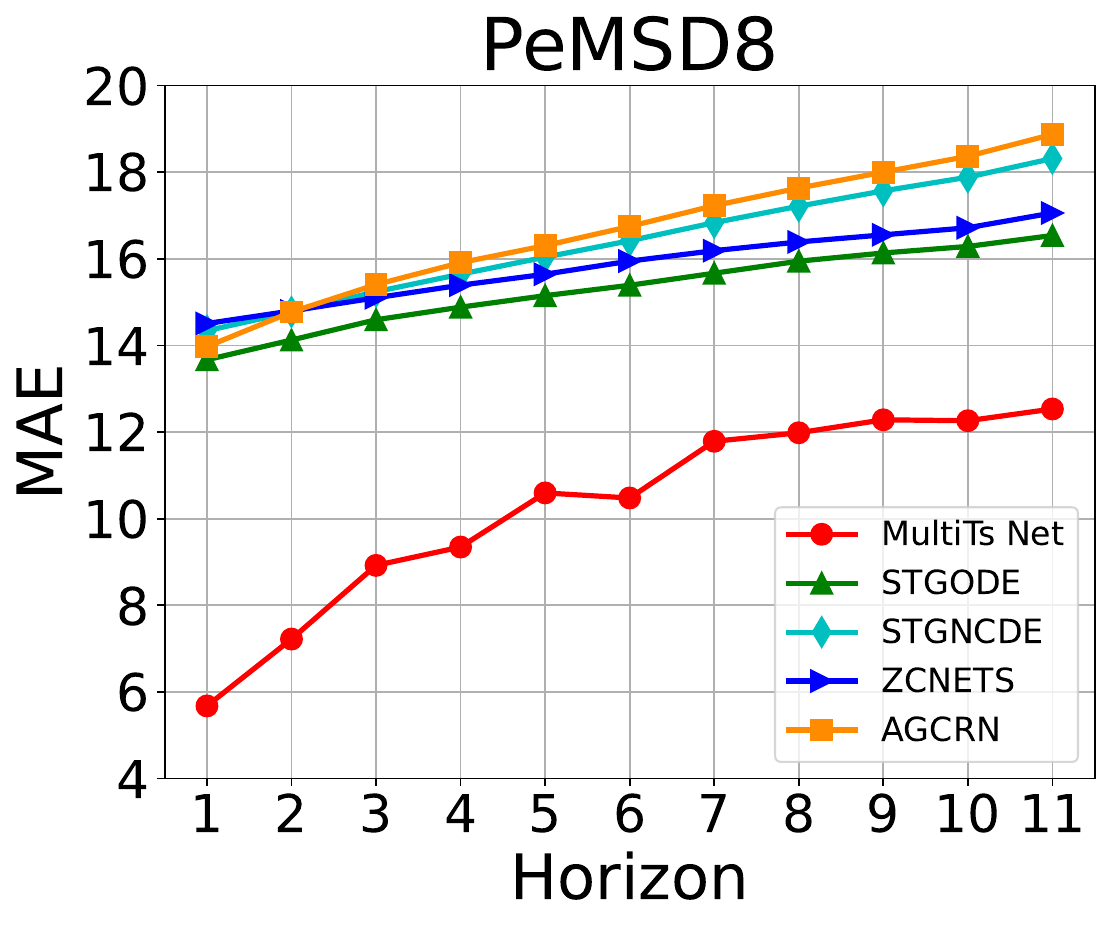}}
\subfloat[MAPE on PeMSD8]{\includegraphics[width=50mm]{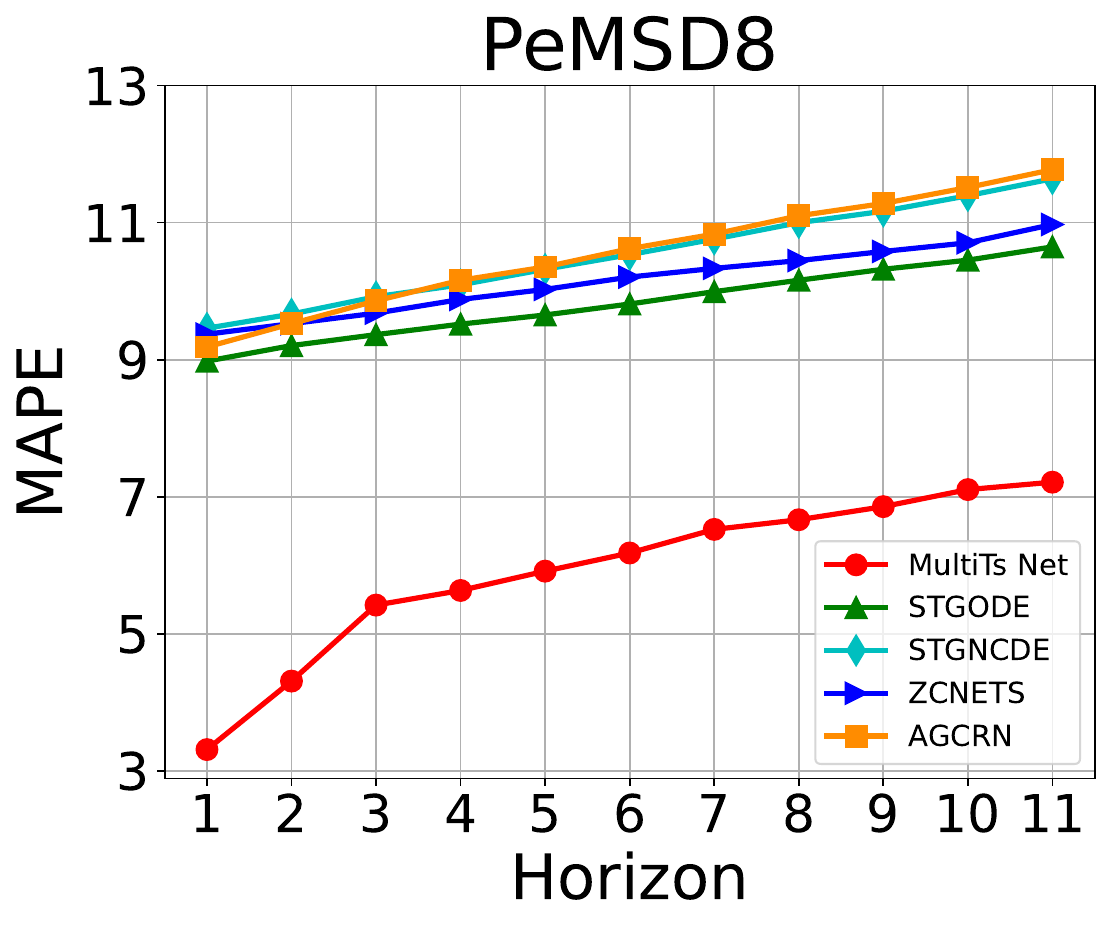}}
\subfloat[RMSE on PeMSD8]{\includegraphics[width=50mm]{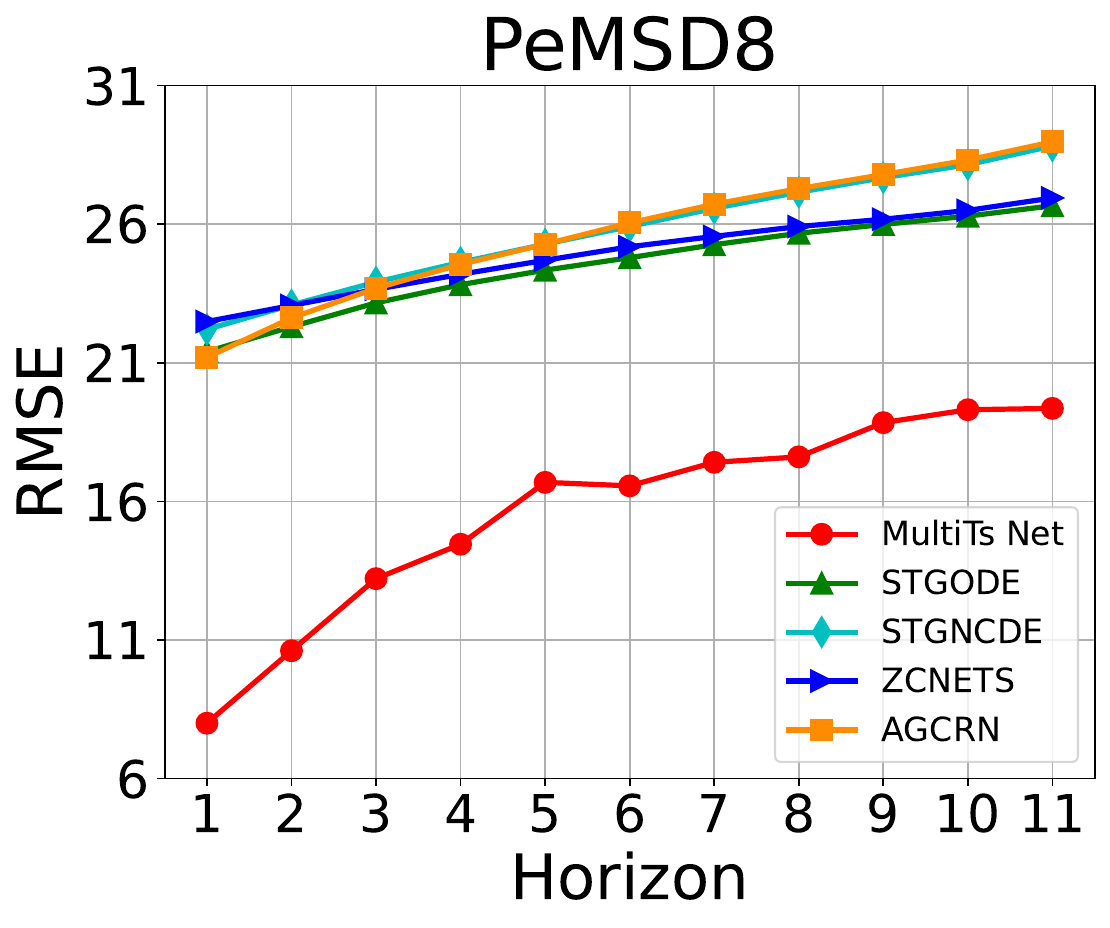}}
}\\[-2ex]
\caption{The table displays the pointwise prediction error across multiple forecast horizons on benchmark datasets.}
\label{fig:ppeh1}
\vspace{-2mm}
\end{figure*}

\begin{figure*}[ht!]
\centering
\hspace*{0cm}\resizebox{0.875\textwidth}{!}{
\subfloat[Node 12 in PeMSD3]{\includegraphics[width=50mm]{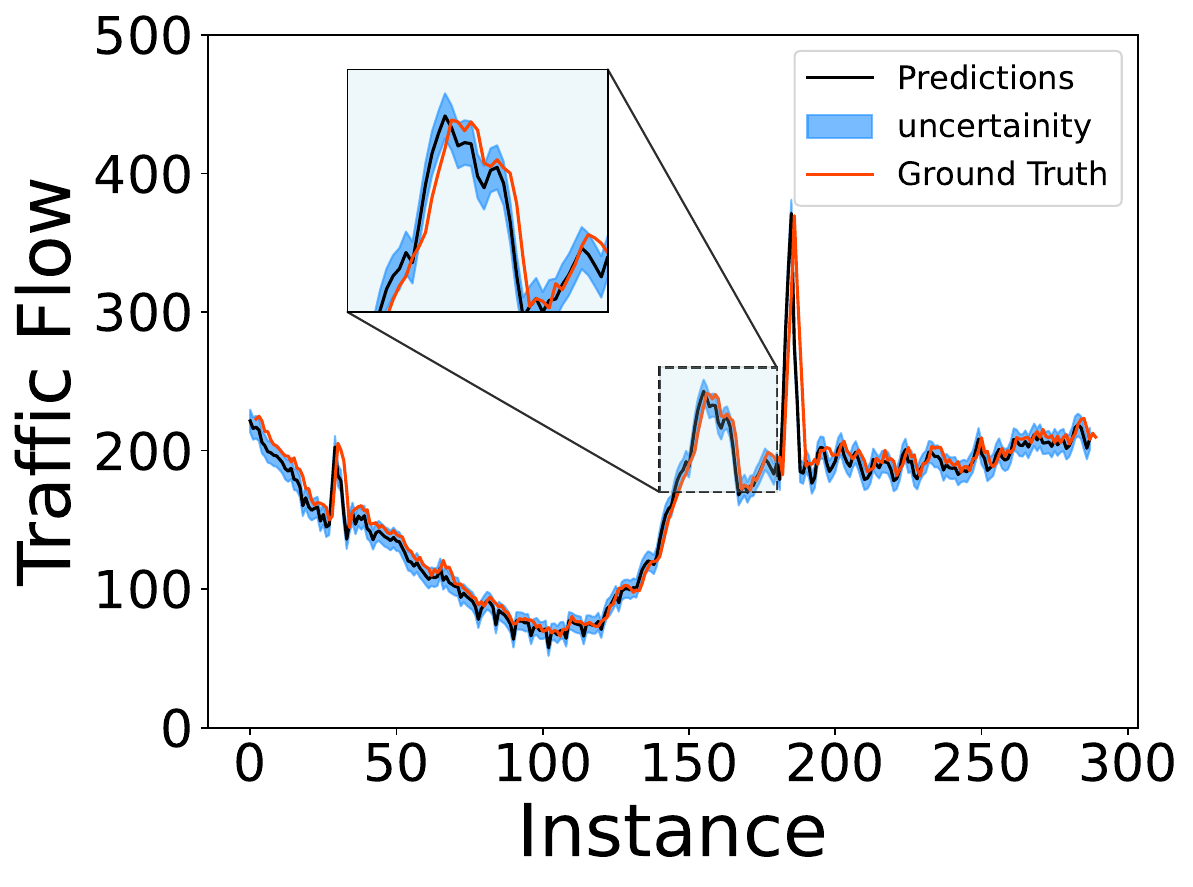}}
\subfloat[Node 99 in PeMSD3]{\includegraphics[width=50mm]{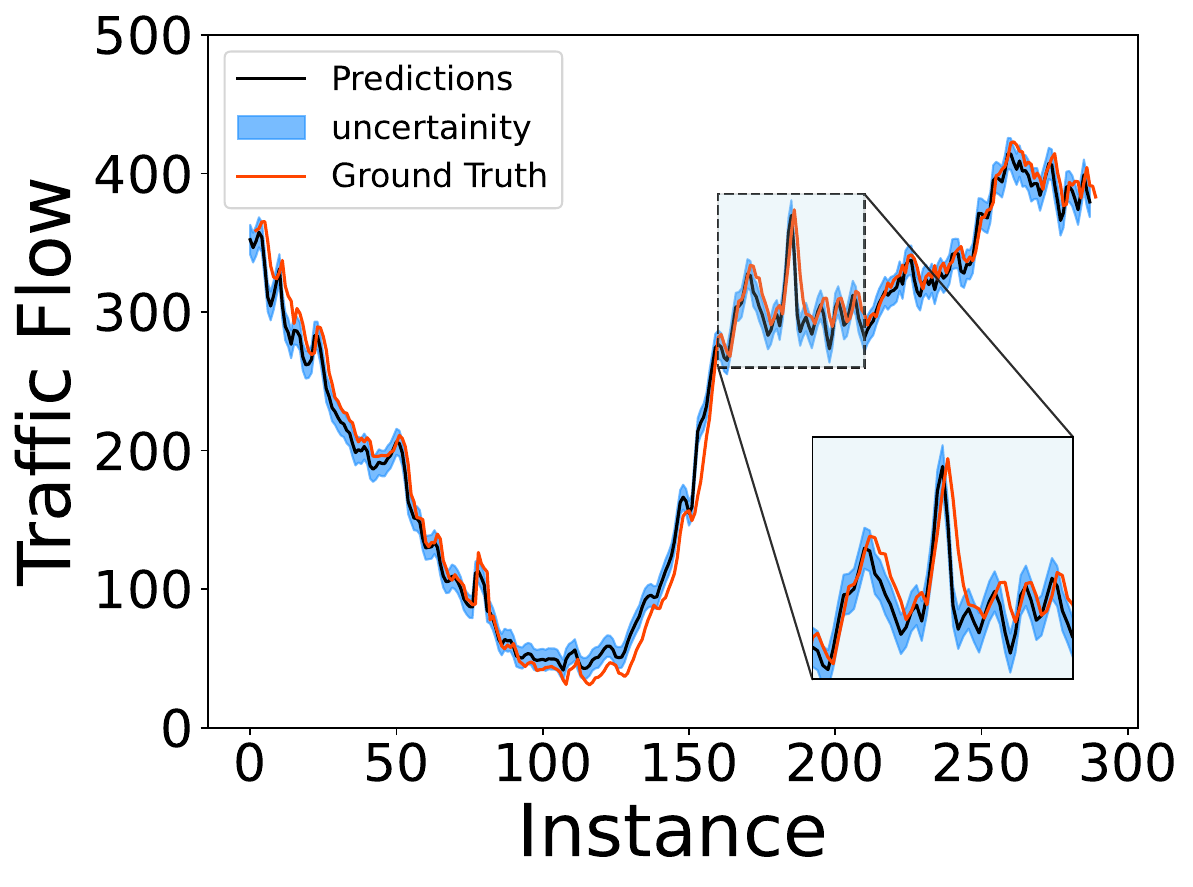}}
\subfloat[Node 108 in PeMSD3]{\includegraphics[width=50mm]{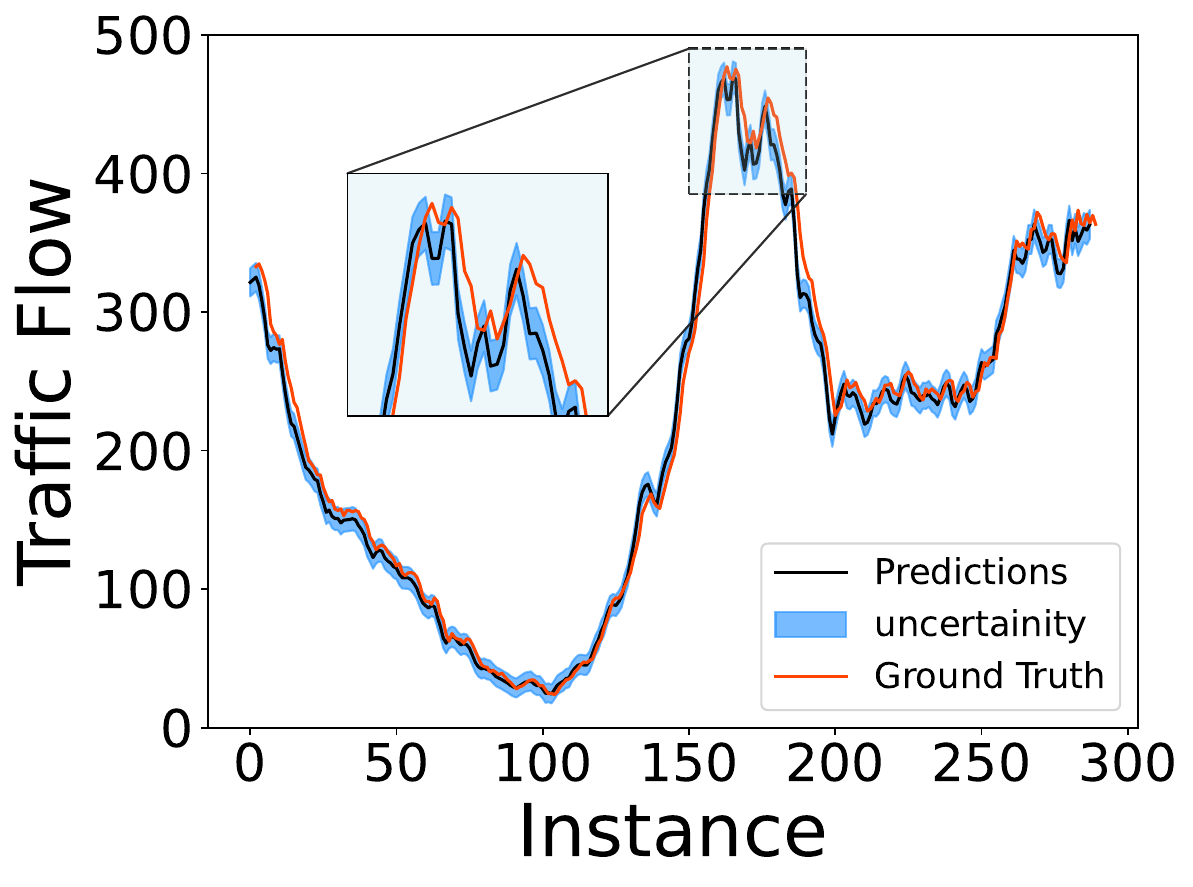}} 
}\\[-2ex]
\hspace*{0cm}\resizebox{0.875\textwidth}{!}{

\subfloat[Node 141 in PeMSD3]{\includegraphics[width=50mm]{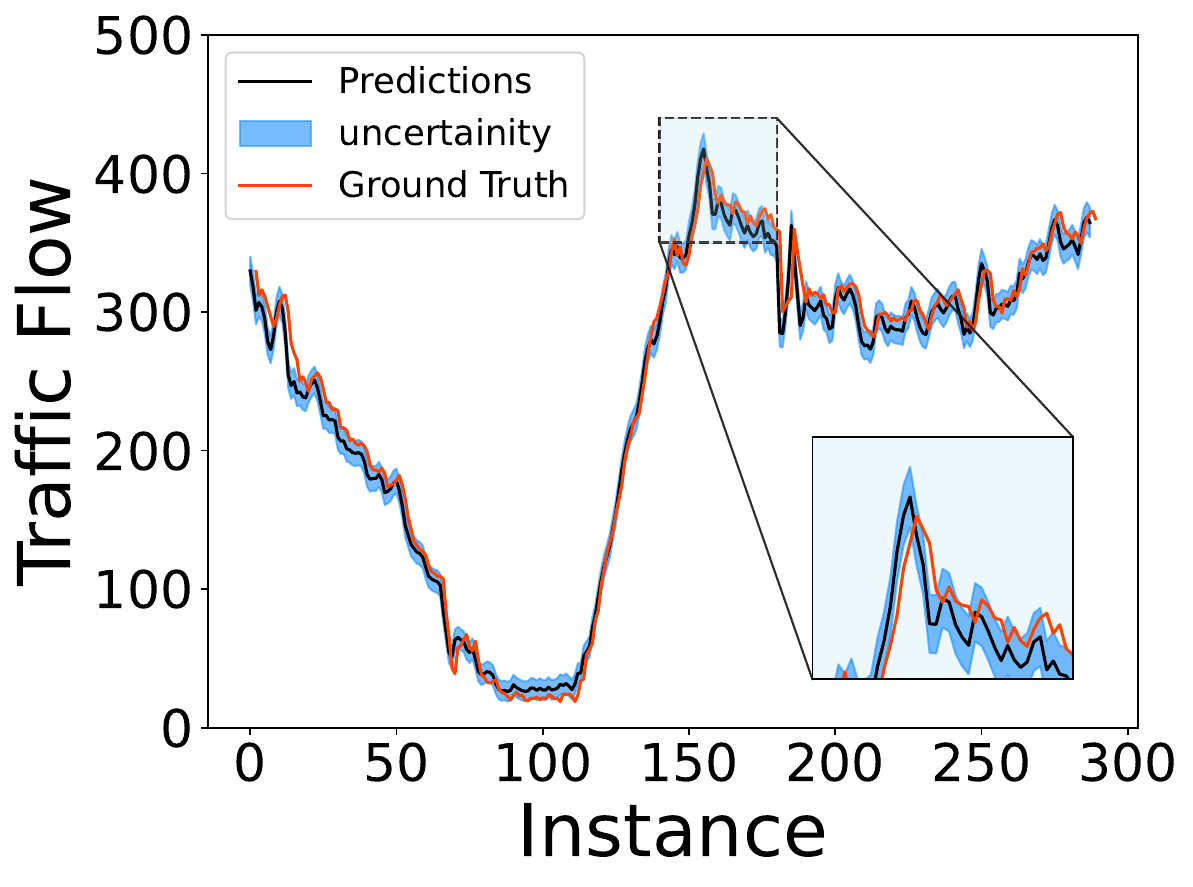}}
\subfloat[Node 149 in PeMSD4]{\includegraphics[width=50mm]{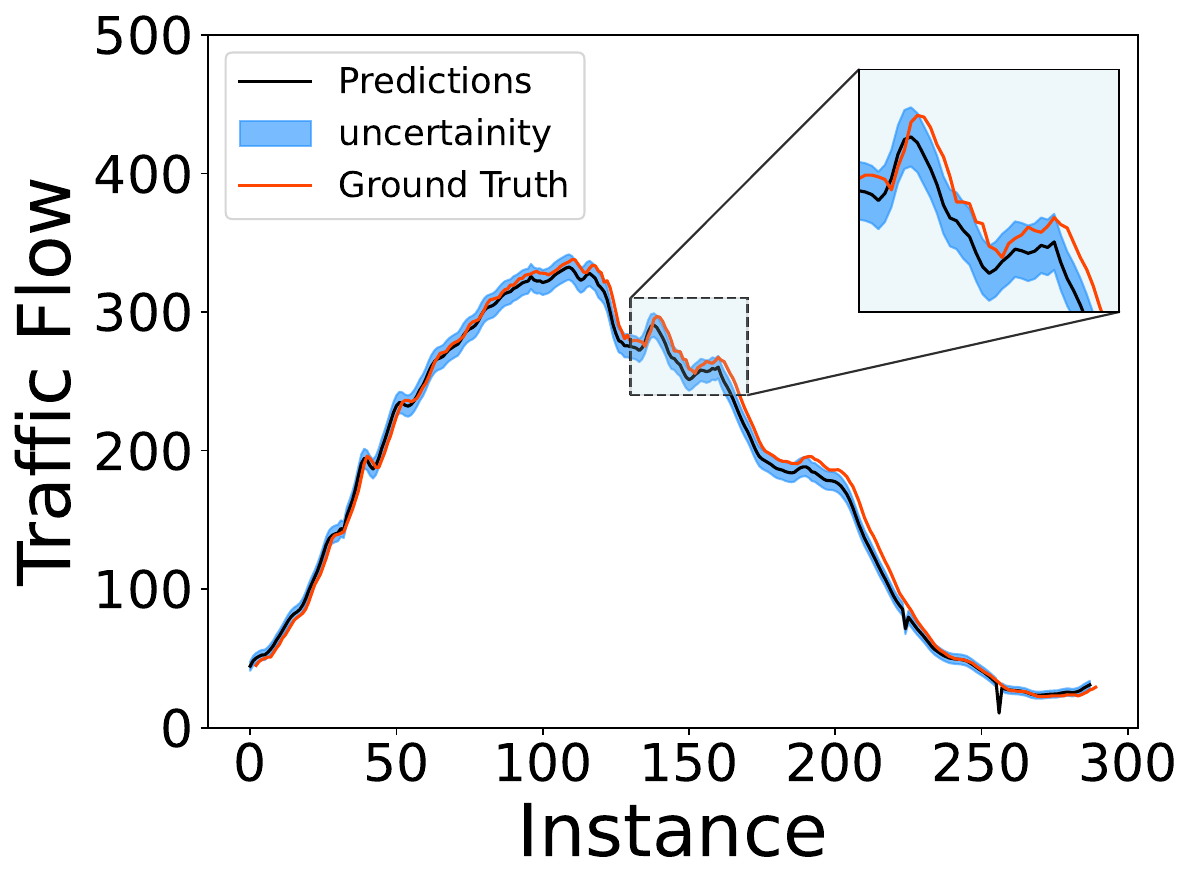}}
\subfloat[Node 170 in PeMSD4]{\includegraphics[width=50mm]{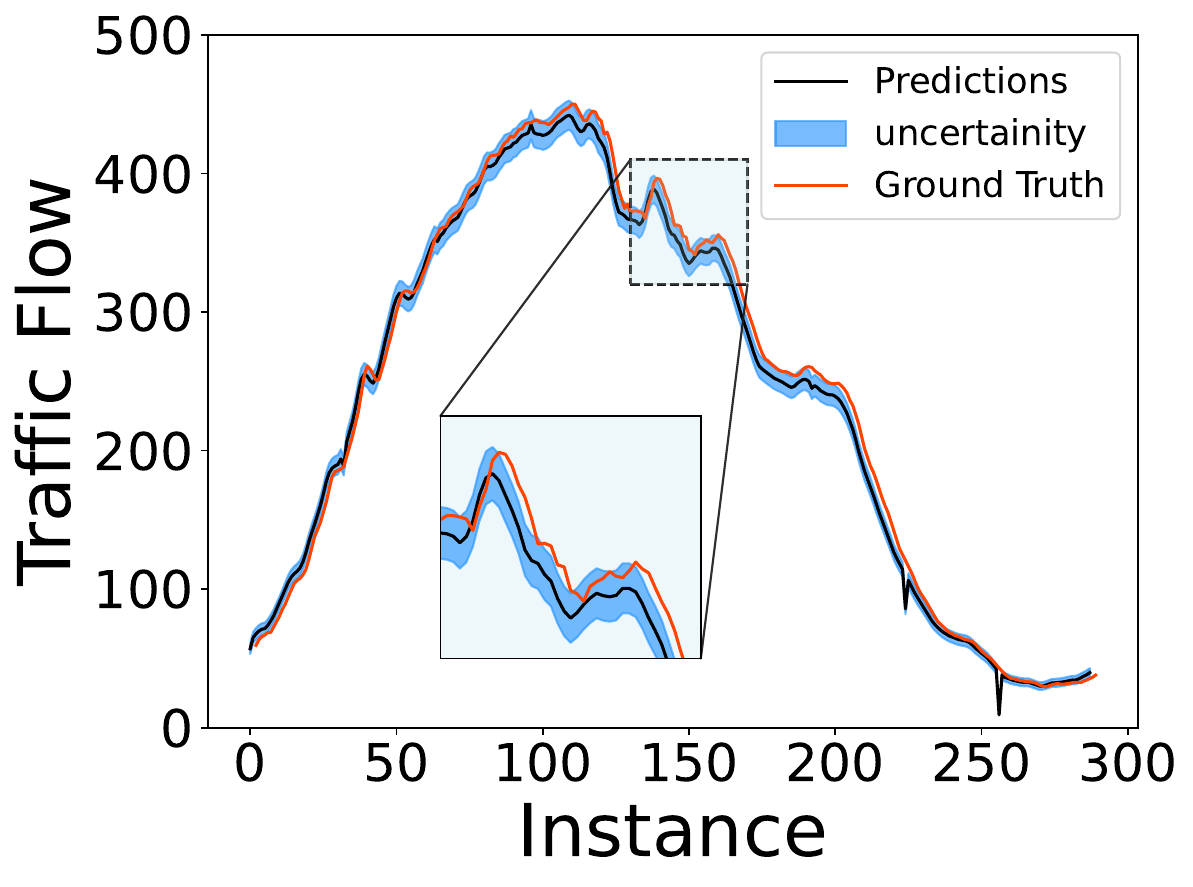}}
}\\[-2ex]
\hspace*{0cm}\resizebox{0.875\textwidth}{!}{
\subfloat[Node 211 in PeMSD4]{\includegraphics[width=50mm]{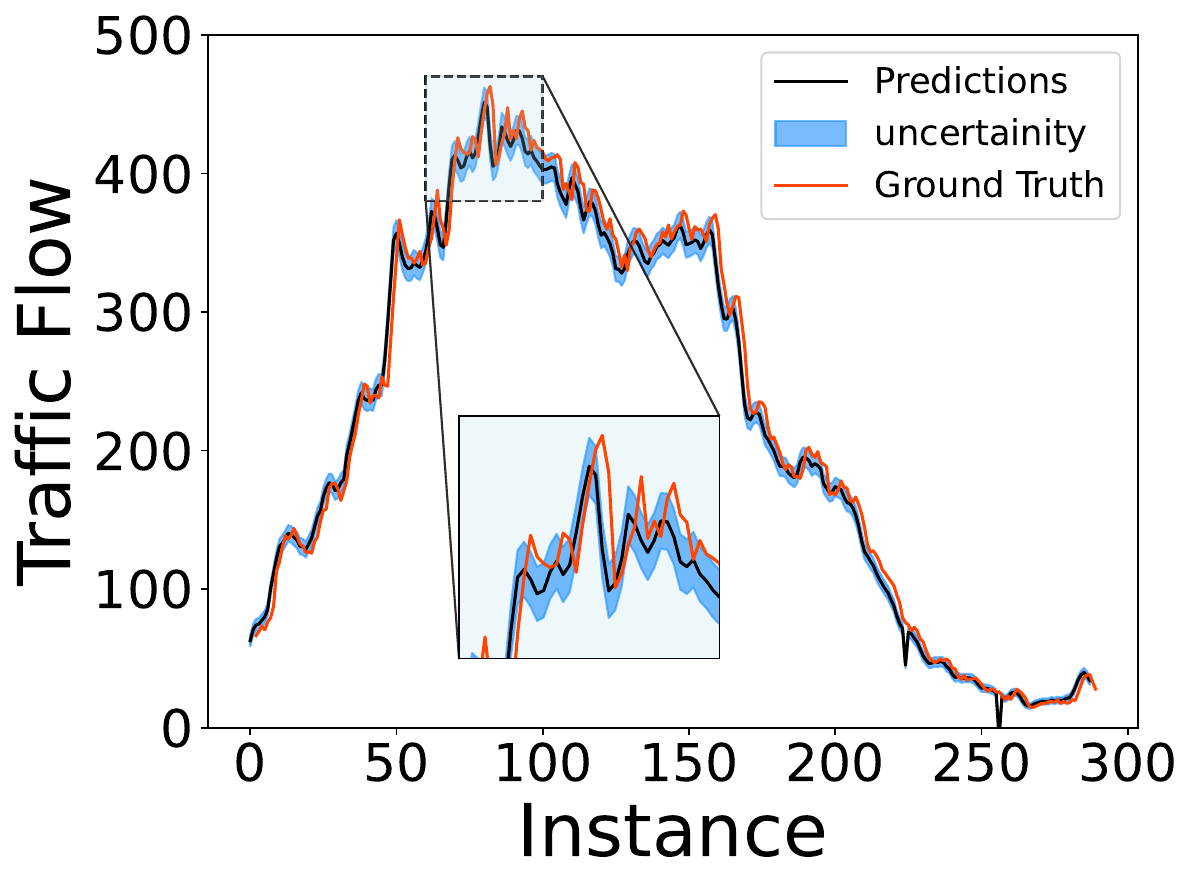}}
\subfloat[Node 287 in PeMSD4]{\includegraphics[width=50mm]{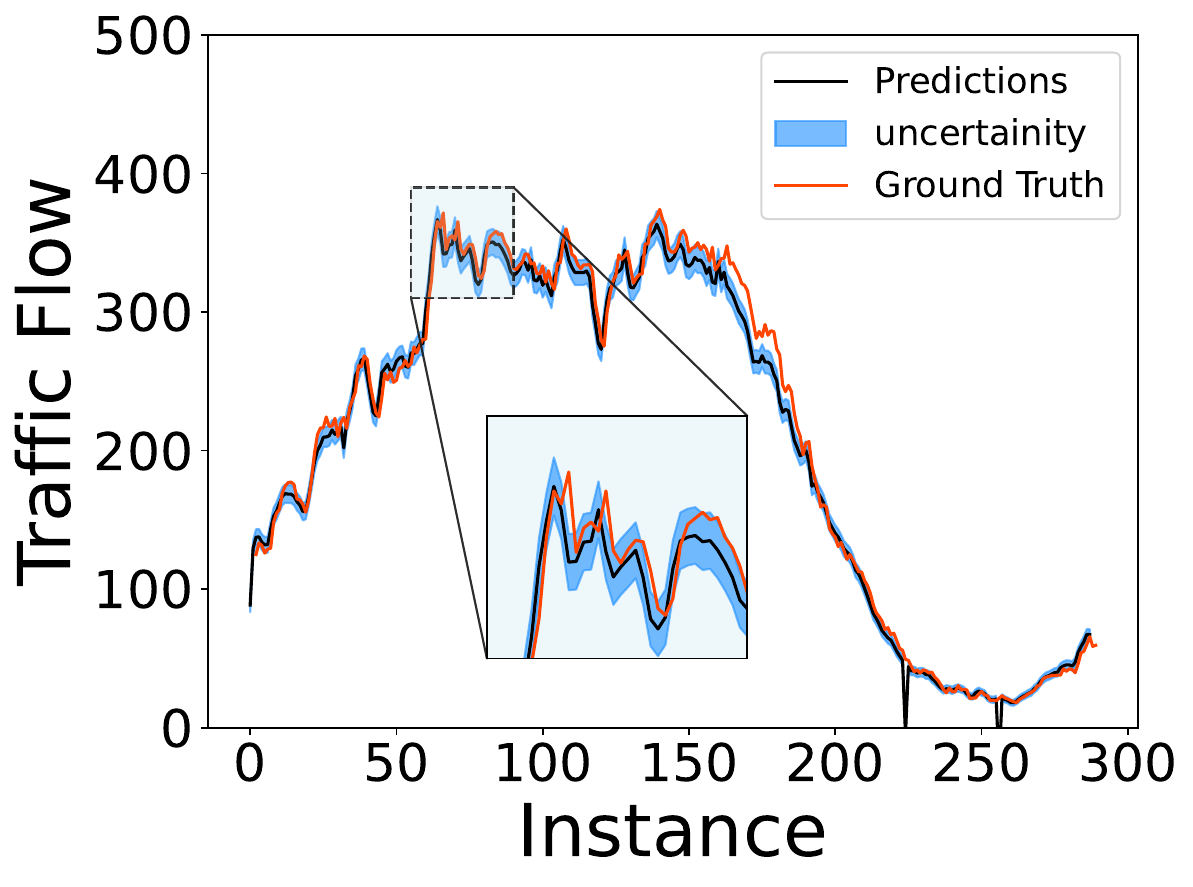}}
\subfloat[Node 66 in PeMSD4]{\includegraphics[width=50mm]{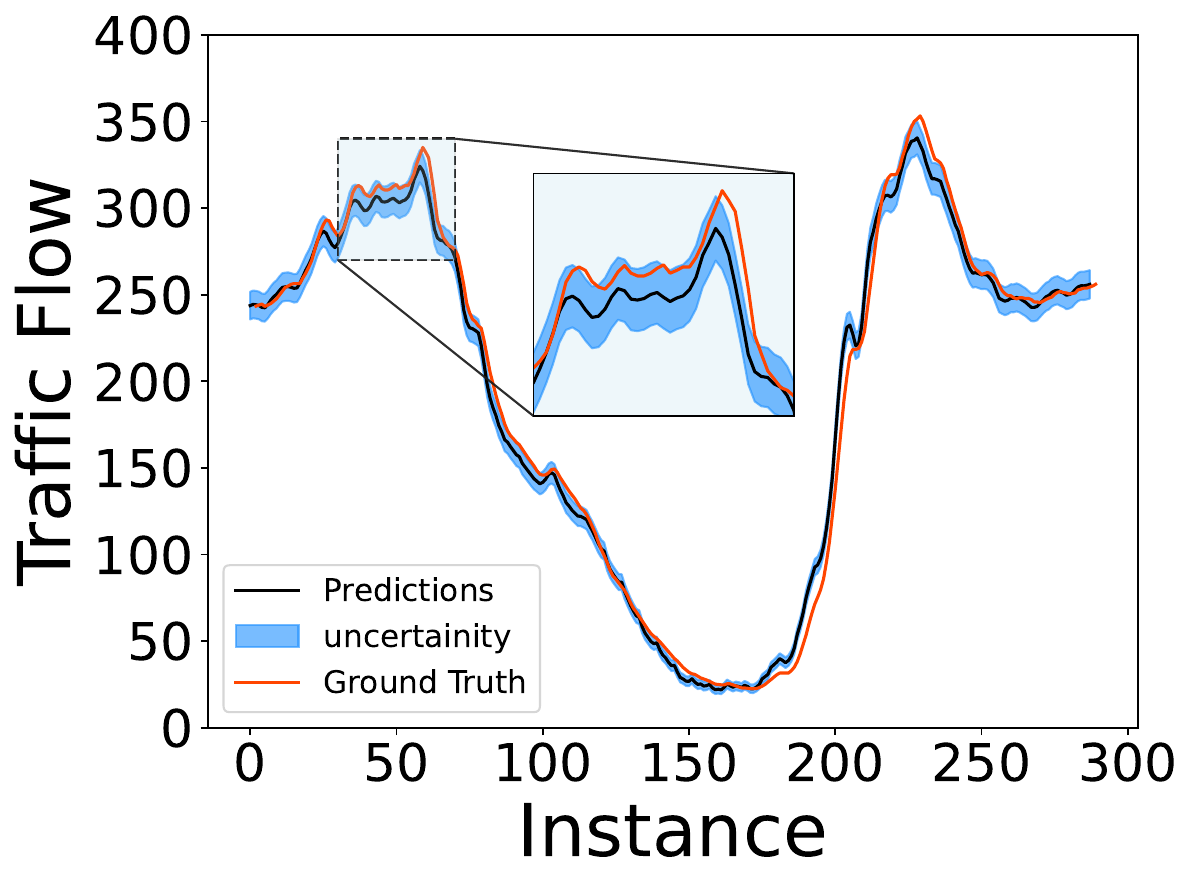}}
}\\[-2ex]
\hspace*{0cm}\resizebox{0.875\textwidth}{!}{
\subfloat[Node 139 in PeMSD4]{\includegraphics[width=50mm]{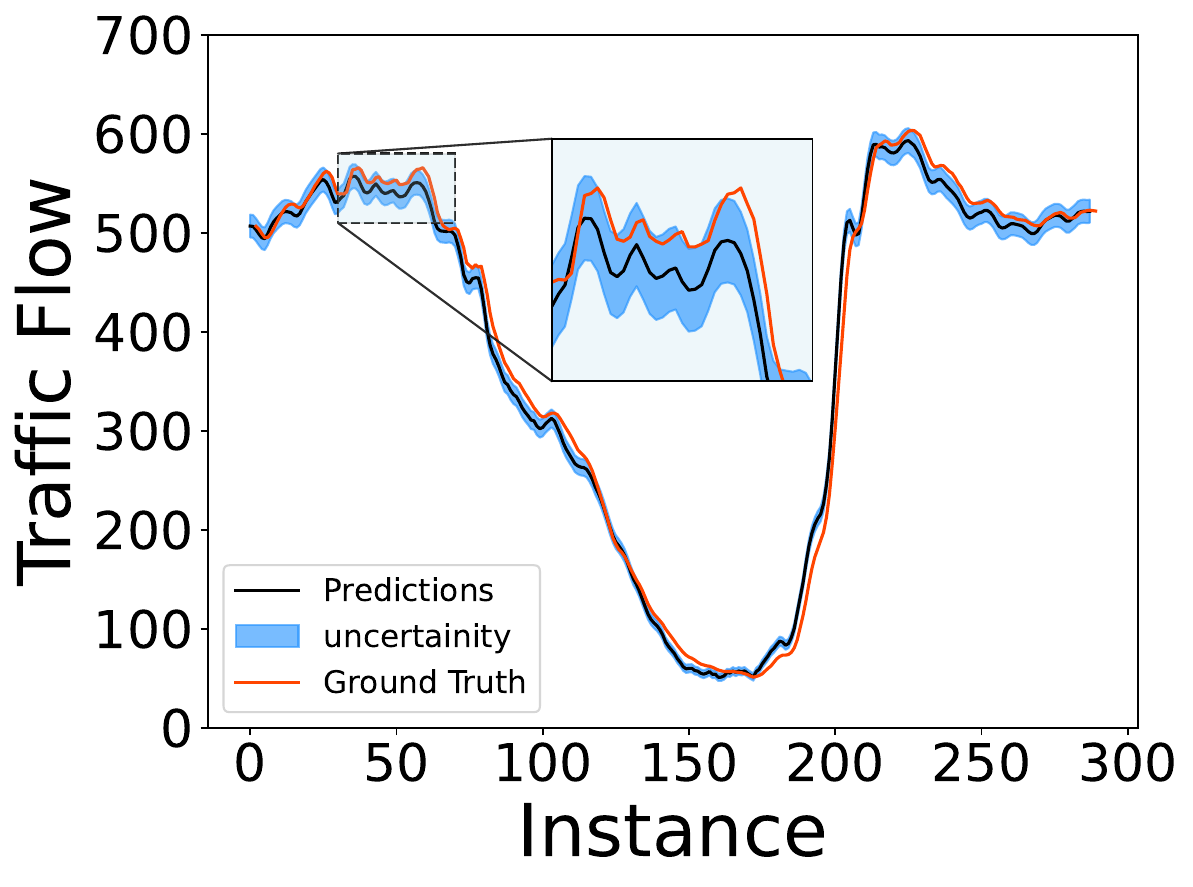}}
\subfloat[Node 277 in PeMSD4]{\includegraphics[width=50mm]{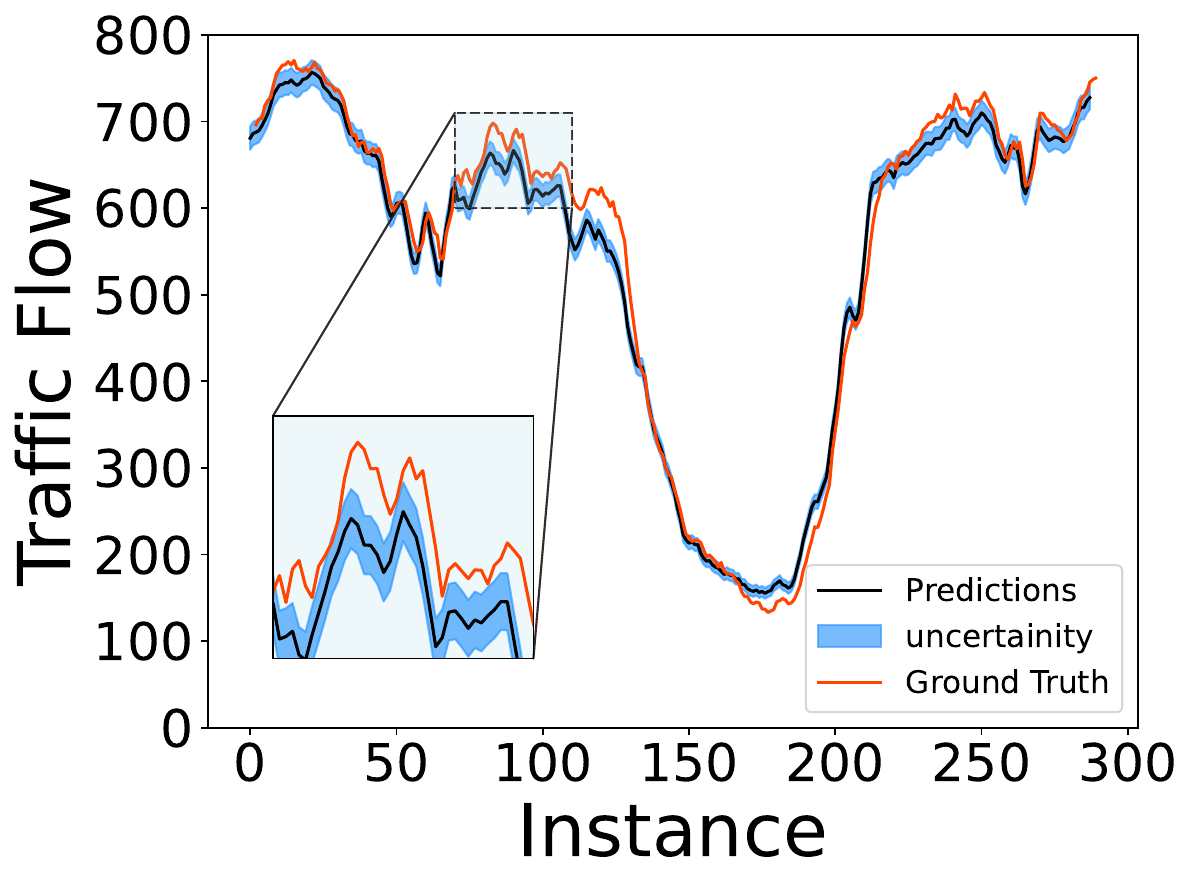}}
\subfloat[Node 85 in PeMSD4]{\includegraphics[width=50mm]{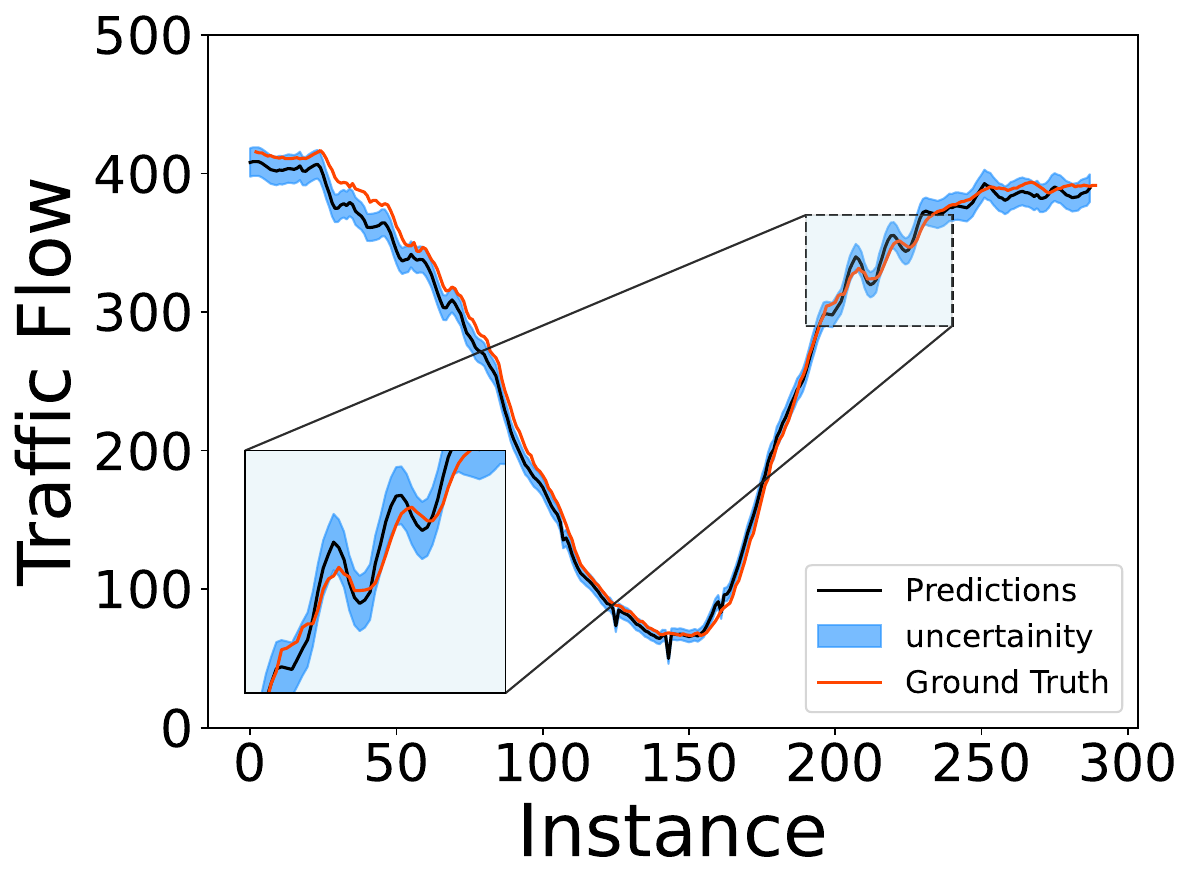}}
}\\[-2ex]
\hspace*{0cm}\resizebox{0.600\textwidth}{!}{
\subfloat[Node 104 in PeMSD8]{\includegraphics[width=50mm]{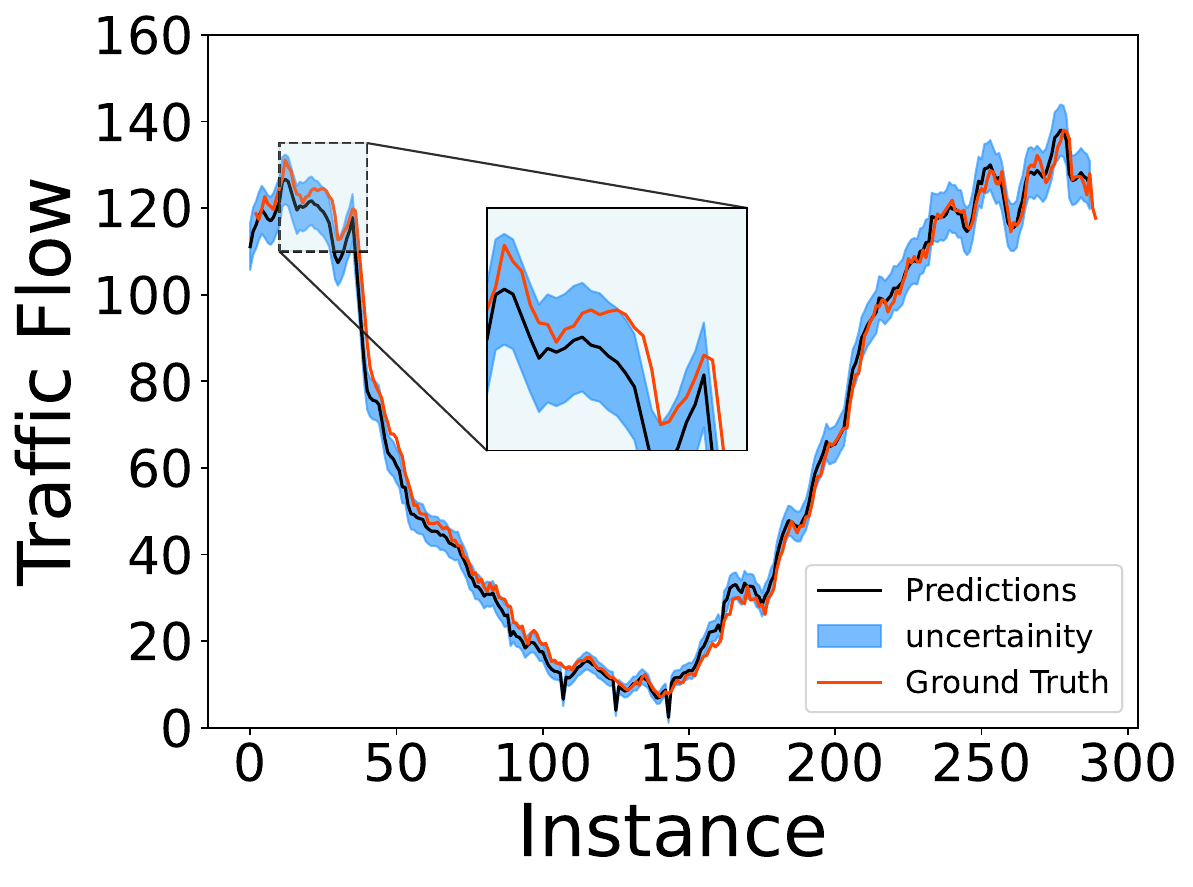}}
\subfloat[Node 155 in PeMSD8]{\includegraphics[width=50mm]{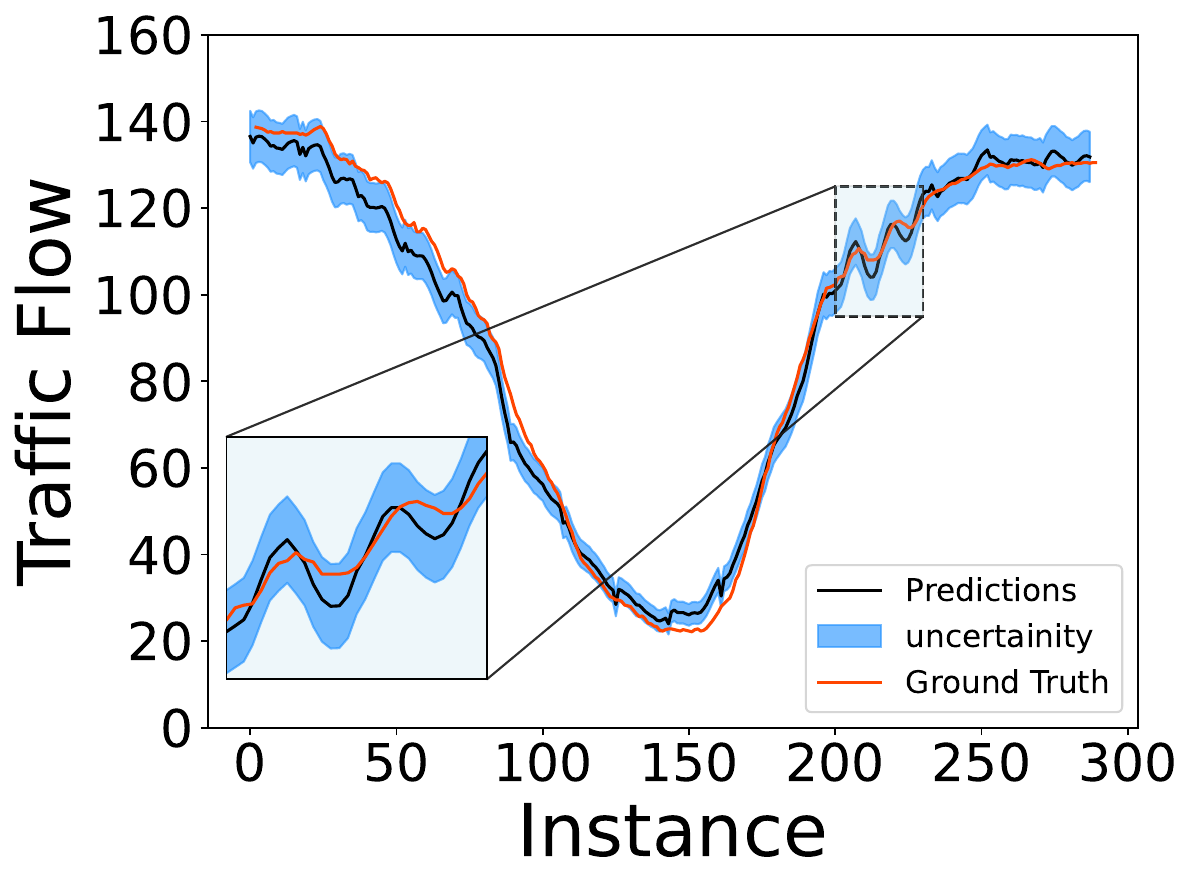}}
}\\[-2ex]
\caption{The figure shows the uncertainty estimations for the \texttt{w/Unc-MultiTs Net} framework forecasts on a sample of sensors(nodes) on the benchmark datasets.}
\label{fig:tfv1}
\end{figure*}

\vspace{-2mm}
Tables \ref{tab:results1} and \ref{tab:results2} compare two models, \texttt{MultiTs} Net and \texttt{w/Unc-MultiTs} Net, against various baselines in the MTSF task. The baseline results are reported from earlier studies\cite{choi2022graph, wu2020connecting}. In a standard benchmark setting, we utilize historical data to predict estimates in a popular 12-sequence-to-12-sequence forecasting task. It involves using 12 time steps of historical data to forecast the value at the 12th future time step and then computing the forecasting errors. Both the \texttt{MultiTs} Net and its variant with local uncertainty estimation, \texttt{w/Unc-MultiTs} Net, demonstrate superior performance over the baselines. They achieve lower forecast errors and effectively capture complex MTS data dynamics. However, \texttt{w/Unc-MultiTs} Net, despite providing uncertainty estimates, slightly underperforms compared to \texttt{MultiTs} Net. Figure \ref{fig:tfv1} shows the uncertainty estimations on framework forecasts on the benchmark datasets. Our framework forecasts consistently outperform baselines, as seen across all prediction horizons in Figure \ref{fig:ppeh1}.

\vspace{-4mm}
\subsection{Ablation Study Results} 
\vspace{-2mm} 
The \texttt{MultiTs} Net is a unified framework designed to improve the accuracy and reliability of forecasting in multi-time series (MTS) data. An ablation study was conducted to evaluate the importance of each component in the framework by testing various ablated variants (with specific components disabled) on multiple datasets. The ablated variants that exclude the language model processing, dynamic prompting mechanism, intra-series dependencies, inter-series dependencies, and multi-modal alignment method are labeled as proposed framework `w/o LLMs', `w/o DP', `w/o IntraS', `w/o InterS', and `w/o MMA' respectively; `w/o' stands for `without'. The study found that removing components significantly reduced performance, highlighting their importance. Table \ref{tab:abstudy} shows the ablation study results. In particular, the ablated variant lacking the multi-modal multi-head attention mechanism (MMA) exhibited the highest error rates, emphasizing its crucial role. The increase in error for the `w/o MMA' ablated variant might be attributed to the oversimplified linear layer substituted to demonstrate MMA effectiveness. Variations in the performance of the ablated variants across different datasets suggest that the complexity of each dataset uniquely affects the effectiveness of each component.

\vspace{-2mm}
\begin{table*}[ht!]
\setlength{\tabcolsep}{0.355em} 
\renewcommand\arraystretch{1.12} 
\centering
 \resizebox{0.925\textwidth}{!}{
\begin{tabular}{c|c|ccc|ccc|ccc}
\hline
\multirow{2}{*}{\textbf{Datasets}}  & \multirow{2}{*}{\textbf{Methods}} & \multicolumn{3}{c|}{\textbf{Horizon$\textbf{@}$3}}       & \multicolumn{3}{c|}{\textbf{Horizon$\textbf{@}$6}}       & \multicolumn{3}{c}{\textbf{Horizon$\textbf{@}$12}}       \\ \cline{3-11} 
                                    &                                   & \textbf{RMSE} & \textbf{MAE}  & \textbf{MAPE} & \textbf{RMSE} & \textbf{MAE}  & \textbf{MAPE} & \textbf{RMSE} & \textbf{MAE}  & \textbf{MAPE} \\ \hline
\multirow{13}{*}{\textbf{METR-LA}}  & HA                       & 10.00         & 4.79          & 11.70         & 11.45         & 5.47          & 13.50         & 13.89         & 6.99          & 17.54         \\
                                    & VAR                      & 7.80          & 4.42          & 13.00         & 9.13          & 5.41          & 12.70         & 10.11         & 6.52          & 15.80         \\
                                    & SVR                      & 8.45          & 3.39          & 9.30          & 10.87         & 5.05          & 12.10         & 13.76         & 6.72          & 16.70         \\
                                    & FC-LSTM                  & 6.30          & 3.44          & 9.60          & 7.23          & 3.77          & 10.09         & 8.69          & 4.37          & 14.00         \\
                                    & DCRNN                    & 5.38          & 2.77          & 7.30          & 6.45          & 3.15          & 8.80          & 7.60          & 3.60          & 10.50         \\
                                    & STGCN                    & 5.74          & 2.88          & 7.62          & 7.24          & 3.47          & 9.57          & 9.40          & 4.59          & 12.70         \\
                                    & Graph WaveNet            & 5.15          & 2.69          & 6.90          & 6.22          & 3.07          & 8.37          & 7.37          & 3.53          & 10.01         \\
                                    & ASTGCN                   & 9.27          & 4.86          & 9.21          & 10.61         & 5.43          & 10.13         & 12.52         & 6.51          & 11.64         \\
                                    & STSGCN                   & 7.62          & 3.31          & 8.06          & 9.77          & 4.13          & 10.29         & 11.66         & 5.06          & 12.91         \\
                                    & MTGNN                    & 5.18          & 2.69          & 6.88          & 6.17          & 3.05          & 8.19          & 7.23          & 3.49          & 9.87          \\
                                    & GMAN                     & 5.55          & 2.80          & 7.41          & 6.49          & 3.12          & 8.73          & 7.35          & 3.44          & 10.07         \\
                                    & DGCRN                    & 5.01          & 2.62          & 6.63          & 6.05 & 2.99 & 8.02          & 7.19 & 3.44 & 9.73 \\ \cline{2-11} 
                                    & \textbf{MultiTs Net}              & \textbf{3.81}  & \textbf{1.89} & \textbf{4.54} & \textbf{5.95} & \textbf{2.94}  & \textbf{7.15} & \textbf{6.92}  & \textbf{3.33} & \textbf{9.47} \\
                                    & \textbf{w/Unc-MultiTs Net}        & \textbf{3.95}  & \textbf{1.96} & \textbf{4.69} & \textbf{6.17} & \textbf{3.17} & \textbf{7.27} & \textbf{6.96} & \textbf{3.42} & \textbf{9.54} \\ \hline                                      
\multirow{13}{*}{\textbf{PEMS-BAY}} & HA                       & 4.30          & 1.89          & 4.16          & 5.82          & 2.50          & 5.62          & 7.54          & 3.31          & 7.65          \\
                                    & VAR                      & 3.16          & 1.74          & 3.60          & 4.25          & 2.32          & 5.00          & 5.44          & 2.93          & 6.50          \\
                                    & SVR                      & 3.59          & 1.85          & 3.80          & 5.18          & 2.48          & 5.50          & 7.08          & 3.28          & 8.00          \\
                                    & FC-LSTM                  & 4.19          & 2.05          & 4.80          & 4.55          & 2.20          & 5.20          & 4.96          & 2.37          & 5.70          \\
                                    & DCRNN                    & 2.95          & 1.38          & 2.90          & 3.97          & 1.74          & 3.90          & 4.74          & 2.07          & 4.90          \\
                                    & STGCN                    & 2.96          & 1.36          & 2.90          & 4.27          & 1.81          & 4.17          & 5.69          & 2.49          & 5.79          \\
                                    & Graph WaveNet            & 2.74          & 1.30          & 2.73          & 3.70          & 1.63          & 3.67          & 4.52          & 1.95          & 4.63          \\
                                    & ASTGCN                   & 3.13          & 1.52          & 3.22          & 4.27          & 2.01          & 4.48          & 5.42          & 2.61          & 6.00          \\
                                    & STSGCN                   & 3.01          & 1.44          & 3.04          & 4.18          & 1.83          & 4.17          & 5.21          & 2.26          & 5.40          \\
                                    & MTGNN                    & 2.79          & 1.32          & 2.77          & 3.74          & 1.65          & 3.69          & 4.49          & 1.94          & 4.53          \\
                                    & GMAN                     & 2.91          & 1.34          & 2.86          & 3.76          & 1.63          & 3.68          & 4.32          & 1.86          & 4.37          \\
                                    & DGCRN                    & 2.69          & 1.28          & 2.66          & 3.63          & 1.59          & 3.55          & 4.42          & 1.89          & 4.43          \\ \cline{2-11} 
                                    & \textbf{MultiTs Net}              & \textbf{1.55} & \textbf{0.78} & \textbf{1.55}  & \textbf{2.36} & \textbf{1.15} & \textbf{2.21}  & \textbf{2.86} & \textbf{1.59} & \textbf{2.91} \\
                                    & \textbf{w/Unc-MultiTs Net}        & \textbf{1.64} & \textbf{0.81} & \textbf{1.56} & \textbf{2.48} & \textbf{1.22}  & \textbf{2.33} & \textbf{3.06}  & \textbf{1.62} & \textbf{3.14} \\ \hline
\end{tabular}
}
\vspace{-3mm}
\caption{The table shows the forecast errors on METR-LA and PEMS-BAY datasets.}
\label{tab:results2}
\vspace{-2mm}
\end{table*}

\vspace{-2mm}
\begin{table*}[ht!]
\centering
\setlength{\tabcolsep}{0.2em} 
\renewcommand\arraystretch{1.14} 
\resizebox{1.075\textwidth}{!}{
\begin{tabular}{c|ccc|ccc|ccc|ccc|ccc}
\hline
\multirow{2}{*}{\textbf{Model}} & \multicolumn{3}{c|}{\textbf{PeMSD3}} & \multicolumn{3}{c|}{\textbf{PeMSD4}} & \multicolumn{3}{c|}{\textbf{PeMSD7}} & \multicolumn{3}{c|}{\textbf{PeMSD8}} & \multicolumn{3}{c}{\textbf{PeMSD7(M)}} \\ \cline{2-16} 
 & \multicolumn{1}{l}{\textbf{MAE}} & \multicolumn{1}{l}{\textbf{RMSE}} & \multicolumn{1}{l|}{\textbf{MAPE}} & \multicolumn{1}{l}{\textbf{MAE}} & \multicolumn{1}{l}{\textbf{RMSE}} & \multicolumn{1}{l|}{\textbf{MAPE}} & \multicolumn{1}{l}{\textbf{MAE}} & \multicolumn{1}{l}{\textbf{RMSE}} & \multicolumn{1}{l|}{\textbf{MAPE}} & \multicolumn{1}{l}{\textbf{MAE}} & \multicolumn{1}{l}{\textbf{RMSE}} & \multicolumn{1}{l|}{\textbf{MAPE}} & \textbf{MAE} & \textbf{RMSE} & \textbf{MAPE} \\ \hline
\textbf{MultiTs Net} & \textbf{11.23} & \textbf{16.93} & \textbf{8.86} & \textbf{14.71} & \textbf{21.11} & \textbf{8.29} & \textbf{17.24} & \textbf{25.53} & \textbf{7.43} & \textbf{12.31} & \textbf{17.91} & \textbf{6.38} & \textbf{2.08} & \textbf{4.13} & \textbf{4.93} \\
\textbf{W/Unc-MultiTs Net} & \textbf{11.44} & \textbf{16.92} & \textbf{9.15} & \textbf{15.26} & \textbf{23.16} & \textbf{8.43} & \textbf{16.88} & \textbf{26.75} & \textbf{8.56} & \textbf{12.42} & \textbf{19.11} & \textbf{6.64} & \textbf{2.08} & \textbf{4.10} & \textbf{5.24} \\ \hline
\textbf{w/o LLMs} & 13.39 & 19.89 & 10.65 & 17.54 & 25.65 & 10.22 & 20.41 & 30.33 & 8.4 & 14.36 & 20.9 & 7.35 & 2.43 & 4.85 & 5.77 \\ \hline
\textbf{w/o DP} & 13.26 & 19.98 & 10.43 & 17.59 & 25.44 & 10.21 & 20.22 & 29.89 & 8.54 & 14.03 & 21.6 & 7.7 & 2.48 & 4.77 & 5.75 \\ \hline
\textbf{w/o IntraS} & 13.18 & 18.41 & 10.36 & 16.38 & 24.33 & 9.52 & 18.94 & 29.29 & 8.03 & 13.6 & 20.45 & 7.38 & 2.37 & 4.48 & 5.54 \\ \hline
\textbf{w/o w/o InterS} & 13.07 & 19.2 & 10.32 & 16.72 & 24.27 & 9.21 & 18.99 & 28.96 & 8.27 & 13.41 & 20.18 & 7.36 & 2.28 & 4.54 & 5.52 \\ \hline
\textbf{w/o CMA} & 15.02 & 22.07 & 11.41 & 19.13 & 27.71 & 10.9 & 21.25 & 33.84 & 9.57 & 15.57 & 22.91 & 8.19 & 2.6 & 5.14 & 6.42 \\ \hline
\end{tabular}
}
\vspace{-2.5mm}
\caption{The table shows the ablation study results on the MTSF task using benchmark datasets. Compared to the original \texttt{MultiTS Net} and \texttt{W/Unc-MultiTS} Net frameworks, the performance of the ablated variants decreases.}
\label{tab:abstudy}
\vspace{-2.5mm}
\end{table*}

\vspace{-4mm}
\section{Conclusion}
\vspace{-4mm}
We propose a dynamic, cost-effective, and privacy-conscious hybrid approach for multi-horizon forecasting, specifically designed for private enterprise adoption. This approach integrates time series trend analysis using instruction-tuning of smaller language models with prompt-augmented, time series representation learning. This combination enables accurate and reliable forecasts, even in the presence of complex inter-variable relationships and non-stationarity. The method leverages a blend of domain expert knowledge and data-driven insights to capture both explicit and implicit variable dependencies, as well as long-range temporal dynamics through a time-then-space modeling approach. It overcomes limitations of traditional forecasting methods, such as fixed historical window lengths, by adapting to the changing nature of time series data and being compatible with consumer-grade hardware. Experiments validate the effectiveness of the hybrid method in improving forecast accuracy and quantifying uncertainty.

\vspace{-4mm}
\section{Technical Appendix}
\vspace{-4mm}
\subsection{Proposed Method}
\label{method}
\vspace{-3mm}
\subsubsection{Mixture of Parameter-Efficient Experts}
\vspace{-3mm}
Low-Rank Adaptation (LoRA) \cite{lora} is a parameter-efficient fine-tuning method for pretrained language models that does not increase inference latency. It enables efficient, task-specific customization by incorporating a set of additional, lightweight trainable parameters into the existing architecture, of pretrained models, without modifying the original pretrained weights. Freezing the original weights helps mitigate catastrophic forgetting by preserving the extensive knowledge acquired by the pretrained models while learning new information. LoRA introduces a pair of low-rank weight matrices, known as adapters, alongside the frozen pretrained weights to capture task-specific information. Specifically, LoRA approximates the weight update of a linear layer as follows:

\vspace{-5mm}
\resizebox{0.9\linewidth}{!}{
\hspace{0mm}\begin{minipage}{\linewidth}
\begin{equation}
Y = (W_0 + \Delta W) X = (W_0 + \alpha BA) X \nonumber
\end{equation}
\end{minipage}
}

\vspace{-1mm}
where $Y \in \mathbb{R}^{\hspace{0.25mm}b \times d_{out}}$ and $X \in \mathbb{R}^{\hspace{0.25mm}b \times d_{in}}$ are the output and input of a linear layer, respectively. $d_{in}$, $d_{out}$ are the input and output dimensions, respectively, and $b$ is the batch size. $W_0 \in \mathbb{R}^{\hspace{0.25mm}d_{in} \times d_{out}}$ is the  pretrained weight matrix, $\Delta W$ is the low-rank approximation of the weight update, and $\alpha$ is a scaling constant. $B \in \mathbb{R}^{\hspace{0.25mm}d_{in} \times r}, A \in \mathbb{R}^{\hspace{0.25mm}r \times d_{out}}$ are projection-down and projection-up weight matrices, respectively. For $d_{in} = d_{out} = d$, the low-rank decomposition technique reduces the number of trainable parameters from $\mathcal{O}(d^2)$ to $\mathcal{O}(2dr)$, where $r \ll d$, thus yielding substantial memory savings. The rank $r$ is a key hyperparameter for effective fine-tuning of large pretrained models using LoRA \cite{lora} for niche tasks, impacting computational complexity and adaptability to new tasks. However, LoRA suffers from high activation memory costs during task-specific adaptation, which are comparable to full-parameter fine-tuning due to the need to store large input activations (or intermediate outputs, like $X \in \mathbb{R}^{\hspace{0.25mm}b \times d_{in}}$) for the computation of gradients of low-rank matrices $B$ and $A$ during backpropagation. Current solutions include selective layer adaptation \cite{lora} or activation recomputation \cite{chen2016training}, but these methods may impact performance. In summary, vanilla LoRA enables efficient LLM adaptation through low-rank weight decomposition but faces challenges related to fine-tuning memory overhead. We further enhance the original LoRA method by reducing the activation memory footprint further, without incurring extra computational costs. We achieve this by freezing the projection-down weight $B$, while redefining the projection-up weight $A$ as the product of a pair of low-rank matrices, $D$ and $C$, where $D \in \mathbb{R}^{r \times \frac{r}{2}}$ remains static, and $C \in \mathbb{R}^{\frac{r}{2} \times d}$ is updated during fine-tuning. This approach reduces trainable parameters and minimizes the size of input activations stored during fine-tuning, which are required for backward propagation during gradient computation, all without adding inference latency. In this approach, the input $X \in \mathbb{R}^{\hspace{0.25mm}b \times d_{in}}$ is initially mapped through $B \in \mathbb{R}^{d_{in} \times r}$ and $D \in \mathbb{R}^{r \times \frac{r}{2}}$ to reduce its dimension to $\frac{r}{2}$, before being projected back up through $C$. This approach significantly reduces the activation memory requirements by limiting the storage of input activation to the output of $X$ transformed by matrix $D$, which is retained from the feed-forward pass to compute the gradient of $C$ during backward propagation. We start with $B$ and $D$ initialized from a normal distribution, and $C$ set to zero, while keeping the adaptation weight matrix $\Delta W = BA = B(DC)$ initially at zero. During fine-tuning, only $C$ is updated, which limits weight updates to a reduced column rank space ($\frac{r}{2}$) defined by the output of $D$. In this work, we propose using a mixture of parameter-efficient experts (MoPEs\cite{zadouri2023pushing}) to synergistically combine the advantages of applying a mixture of experts (MoEs) to parameter-efficient fine-tuning (PEFT) methods. This results in a parameter-efficient adaptation of the MoE approach. The LoRA variant achieves parameter efficiency and activation memory reduction, and the MoE architecture utilizes specialized experts (multiple LoRA variants) tailored for adapting to distinct aspects of the input data. With a mixture of experts, each targeting specific patterns in the input data, MoPEs enable more efficient fine-tuning of large pretrained models and enhance overall performance on complex tasks. MoPEs represent a family of neural network architectures that enable conditional computation through multiple experts (LoRA variants), activated based on a gating mechanism (router $R$). We denote the set of $K$ experts as $\{C_{0} = E(X; \theta_{0}), \ldots, C_{K} = E(X; \theta_{K})\}$, where $C_{k}$ is the $k$-th expert weight matrix, learned during fine-tuning. Here, $E$ represents a parameterized function, and $\theta_{k}$ denotes the trainable parameters of expert $k$. The router $R$ is typically another feed-forward network that produces a $k$-dimensional vector indicating the routing probabilities for each expert.  The output of linear layer is computed as follows,

\vspace{-2mm}
\resizebox{0.90\linewidth}{!}{
\hspace{0mm}\begin{minipage}{\linewidth}
\begin{equation}
Y = (W_0 + \Delta W) X = W_{0} X + BD(\overline{C}X), \overline{C}=\sum_{k=1}^{K} R(X)_{k} C_{k} \nonumber
\end{equation}
\end{minipage}
}

\vspace{-2mm}
Here, $\overline{C}$ represents a composite weight matrix obtained by combining the contributions of multiple expert weight matrices, each weighted by its respective routing probability. We employ a top-$k$ routing strategy for soft merging, selecting only the top-$k$ experts for computing the composite matrix, thereby reducing computational complexity. Despite the conditional computation facilitated by the MoPEs approach, the reduction in activation memory allows for an affordable fine-tuning approach on consumer hardware.

\vspace{-4mm}
\subsubsection{Fine-Tuning small-scale LMs}
\vspace{-2mm}
The Llama 2\cite{touvron2023llama} is an advanced autoregressive, language-optimized transformer architecture, fine-tuned using supervised fine-tuning (SFT) and reinforcement learning with human feedback (RLHF) to align with human-centric values and preferences. It incorporates RMSNorm pre-normalization, PaLM-inspired SwiGLU activation functions, and rotary positional embeddings. Additionally, it utilizes a grouped-query attention mechanism, extending the input context length to 4096 tokens. The architecture consists of 32 layers and 32 attention heads, with a hidden size of 4096, and supports batch sizes of up to 32 for sequences up to 2048 tokens. We utilize zero-shot prompting of the Llama2-70B model to generate training data for time series trend analysis, enabling task-specific fine-tuning of smaller models. We perform instruction tuning on smaller models, such as Llama2-7B through the Quantized MoPEs technique, using the machine-generated data mentioned earlier, to efficiently customize them for niche time series trend analysis through transfer learning. This approach allows us to achieve both accuracy and efficiency comparable to that of the larger model. We have integrated MoPEs modules into each linear layer of the grouped-query attention layers in the Llama2-7B model architecture for efficient fine-tuning. Each layer typically captures different aspects of language, with lower layers often capturing basic syntactic information and higher layers capturing more complex semantic relationships, allowing for task-specific adaptation. Furthermore, the original weights of the Llama2-7B model hosted by Meta AI are in 16-bit format to reduce memory usage. We also apply 8-bit quantization\cite{dettmers2023qlora} to further compress the pretrained language model's parameters, significantly reducing memory and computational costs. We leverage paired input time series data and their corresponding Llama2-70B-generated textual summaries to instruct-tune a smaller Llama2-7B model and minimize standard cross-entropy loss to achieve similar performance with reduced resource consumption and increased interpretability. The Llama2-7B model compute expressive token embeddings to encapsulate both contextual information and semantic relationships between words or phrases.  We freeze the fine-tuned Llama2-7B model and use a downstream, forecasting task-based, differentiable softmax attention pooling mechanism to derive text-level embeddings, represented as $H_{\text{text}} \in \mathbb{R}^{N \times W \times d}$, across a historical time window to compliment traditional forecasting method. Our innovative method aims to demystify the `black-box' nature of the Llama2-70B by generating instruction-following data, thereby enhancing the Llama2-7B's capabilities in interpreting and analyzing time series data with increased precision and explainability through task-specific customization.

\vspace{-4mm}
\subsubsection{Dynamic Prompting Mechanism Design}
\vspace{-2mm} 
We present a dynamic prompting mechanism designed to enhance the adaptability and accuracy of traditional forecasting methods when dealing with complex time series data. The dynamic prompting mechanism consists of a predefined set of shared pools of prompts, stored as key-value pairs, with each prompt associated with specific time series characteristics, such as periodic trends, seasonality, cyclicality, and more. The prompting mechanism enables traditional methods to retrieve relevant prompts based on the evolving nature of time series data and apply learned patterns for forecasting tasks. This allows them to draw upon appropriate past knowledge and adapt to new, similar time series trends or patterns, ultimately leading to improved forecast accuracy. Traditional methods often struggle to adapt to dynamic, non-stationary data with distributional shifts. The ability to access and utilize the most relevant prompts from the shared pool to introduce appropriate time-series-specific prior knowledge significantly improves upon traditional methods. The shared pool of prompts encodes contextual information and insights learned from historical time series data stored as key-value pairs \( (\boldsymbol{k}_m, V_m) \) described as follows:   

\vspace{0mm}
\resizebox{0.93\linewidth}{!}{
\hspace{0mm}\begin{minipage}{\linewidth}
\begin{equation}
    \mathbf{P}=\left\{\left(\boldsymbol{k}_1, V_1\right),\left(\boldsymbol{k}_2, V_2\right), \cdots,\left(\boldsymbol{k}_M, V_M\right)\right\} \nonumber
\end{equation}
\end{minipage}
}

\vspace{0mm}
Where \( M \) represents the total number of prompts in this robust and representative set. Within this set, denoted as \( \mathbf{K} = \{ \boldsymbol{k}_m \}_{m=1}^M \), the key for each prompt is \( \boldsymbol{k}_m \in \mathbb{R}^{d} \). Correspondingly, in the set \( \mathbf{V} = \{ V_m \}_{m=1}^M \), each \( V_m \in \mathbb{R}^{W \times d} \) represents the value of the prompt, characterized by a token length \( W \) and an embedding dimension \( d \).  The initial time series data from each sensor, denoted as \(\mathbf{S}^{t}_{i} \in \mathbb{R}^{W}\), is projected through a shared linear layer to $d$-dimensional space, resulting in transformed data represented as \(\mathbf{S}^{t}_{i} \in \mathbb{R}^{W \times d}\). 
The proposed method utilizes an additive attention mechanism to calculate the relevance of each prompt to the current input time series data, enabling the selection of the most pertinent prompts for dynamic adaptation. This approach ensures that the framework can handle a wide spectrum of time series patterns, including those not encountered during training, leading to more accurate predictions. For the additive attention mechanism, the relevance score between the time series data \( \mathbf{S}^{t}_{i} \) and a prompt key \( \boldsymbol{k}_m \) is calculated using a score-matching function computed as follows:

\vspace{-2mm}
\resizebox{0.93\linewidth}{!}{
\hspace{0mm}\begin{minipage}{\linewidth}
\begin{equation}
    a(\mathbf{S}^{t}_{i}, \boldsymbol{k}_m) = \mathbf{W}_v^\top \text{tanh}(\mathbf{W}_q \mathbf{S}^{t}_{i} + \mathbf{W}_k \boldsymbol{k}_m) \nonumber
\end{equation}
\end{minipage}
}

where \( \mathbf{W}_q \in \mathbb{R}^{W \times d} \) is the query weight matrix, \( \mathbf{W}_k \in \mathbb{R}^{d \times d} \) is the key weight matrix, \( \mathbf{W}_v \in \mathbb{R}^{d} \) is the trainable vector for computing the final score, and \( a: \mathbb{R}^{W \times d} \times \mathbb{R}^{d} \rightarrow \mathbb{R} \) is the function mapping the input data and key to a relevance score. The score-matching function identifies the most similar prompt keys and selects the corresponding prompt values, optimizing forecasts with improved accuracy based on the retrieved prompts. The relevance score measures the significance of a specific prompt for the current time series data, aiding the framework in identifying the most relevant past knowledge. We retrieve the top-\( \mathbf{K} \) corresponding prompt values for the input data \( \mathbf{S}^{t}_{i} \) and concatenate them to obtain the contextualized time series embedding for each sensor: \( \mathbf{\tilde{S}}^{t}_{i} = \left[V_{1} ; \cdots ; V_{\mathbf{K}} ; \mathbf{S}^{t}_{i}\right]\mathbf{W}_{o} \), where \( \mathbf{W}_{o} \) is trainable weight matrix. Employing a diverse prompt pool capturing diverse time series traits enables the forecasting framework to adapt to evolving relationships, boosting accuracy and understanding of complex patterns. In the coming sections, we effectively model both intra-series and inter-series dependencies in MTS data, leading to more accurate and robust forecasting models.

\vspace{-3mm}
\subsubsection{Modeling Intra-series dependencies} 
\vspace{-2mm} 
We model the dependencies within each individual time series to enhance pointwise forecasts. We employ the Grouped-query multi-head attention (GQ-MHA) mechanism to capture non-linear, time-evolving dependencies. Our approach involves projecting the time series embedding $\mathbf{\tilde{S}}^{t} \in \mathbb{R}^{N \times W \times d}$ for each of the $N$ sensors into shared keys ($K_g$), shared values ($V_g$), and unique queries ($Q_{g,h}$) for each head(h) in the group(g), as follows:

\vspace{-3mm}
\resizebox{0.89\linewidth}{!}{
\begin{minipage}{\linewidth}
\begin{align*}
   K_g^i &= \mathbf{\tilde{S}}_i^{t} W_{K_g}, \quad i = 1, \ldots, N \\
   V_g^i &= \mathbf{\tilde{S}}_i^{t} W_{V_g}, \quad i = 1, \ldots, N \\
   Q_{g,h}^i &= \mathbf{\tilde{S}}_i^{t} W_{Q_{g,h}}, \quad i = 1, \ldots, N.
\end{align*}
 \end{minipage}
}

\vspace{1mm}
where the weight matrices $W_{K_g}$, $W_{V_g}$, and $W_{Q_{g,h}}$ have dimensions $\mathbb{R}^{d \times d}$. Consequently, the dimensions of $K_g^i$, $V_g^i$, and $Q_{g,h}^i$ for each sensor $i$ are $\mathbb{R}^{W \times d}$, respectively. The transformed time series embeddings are computed using the scaled dot-product attention mechanism, as follows:
 
 \vspace{-1mm}
\resizebox{0.89\linewidth}{!}{
\begin{minipage}{\linewidth}
\begin{equation}
\text{Attention}(Q_{g,h}^i, K_g^i, V_g^i) = \text{softmax}\left( \frac{Q_{g,h}^i (K_g^i)^T}{\sqrt{d_k}} \right) V_g^i \nonumber
\end{equation}
 \end{minipage}
}  

where $d_k = \frac{d}{H}$ is a scaling factor, and $H$ is the total number of heads. We then perform aggregation across heads and groups to synthesize a concise representation of the time series data for each sensor as follows:

\vspace{-3mm}
\resizebox{0.89\linewidth}{!}{
\begin{minipage}{\linewidth}
\begin{equation}
\mathbf{\tilde{S}}_{i}^{t} = \frac{1}{G} \sum_{g=1}^G(\text{Concat}(\text{Attention}_1, \ldots, \text{Attention}_H) W_o) \nonumber
\end{equation}
 \end{minipage}
}

Here, $W_o \in \mathbb{R}^{Hd \times d}$. Understanding the internal dynamics of each time series can provide a strong foundation for later exploring inter-series dependencies.

\vspace{-3mm}
\subsubsection{Modeling Inter-series dependencies} 
\vspace{-2mm}
In highly intricate multivariate systems with intertwined dynamics, a hybrid approach that iteratively learns both intra-series and inter-series dependencies might be the most effective way to adequately capture the full complexity of multivariate data. We perform multi-task learning by jointly modeling intra-series patterns and shared dependency patterns among multiple related time series to enhance predictions. We utilize the Grouped-query multi-head attention mechanism (GQ-MHA) to learn and understand the interdependencies among time series, aiming to enhance pointwise forecasts. Our approach involves projecting the contextualized time series embedding, $\mathbf{\tilde{S}}^{t} \in \mathbb{R}^{N \times W \times d}$, to compute the shared keys $K_g^w$, shared values $V_g^w$, and unique query projections $Q_{g,h}^w$ for each window step ($w$) and group ($g$) as follows:

\vspace{-2mm}
\resizebox{0.89\linewidth}{!}{
\begin{minipage}{\linewidth}
\begin{align*}
   K_g^w &= \mathbf{\tilde{S}}_w^{t} W_{K_g}, \quad w = 1, \ldots, W \\
   V_g^w &= \mathbf{\tilde{S}}_w^{t} W_{V_g}, \quad w = 1, \ldots, W \\
   Q_{g,h}^w &= \mathbf{\tilde{S}}_w^{t} W_{Q_{g,h}}, \quad w = 1, \ldots, W.
\end{align*}
\end{minipage}
}

\vspace{1mm}
Here, $W_{K_g}$, $W_{V_g}$, and $W_{Q_{g,h}}$ are weight matrices with dimensions $\mathbb{R}^{\hspace{0.25mm}d \times d}$. The spatial attention mechanism transformed time series embeddings are computed using a scaled dot-product attention mechanism, as follows,

\vspace{-3mm}
\resizebox{0.89\linewidth}{!}{
\begin{minipage}{\linewidth}
\begin{equation}
\text{Attention}(Q_{g,h}^w, K_g^w, V_g^w) = \text{softmax}\left( \frac{Q_{g,h}^w (K_g^w)^T}{\sqrt{d_k}} \right) V_g^w \nonumber
\end{equation}
\end{minipage}
}

We aggregate the attention scores across all heads and groups to synthesize a comprehensive representation as,
 
\vspace{-3mm}
\resizebox{0.89\linewidth}{!}{
\begin{minipage}{\linewidth}
\begin{equation}
\hspace{-3mm}\mathbf{\bar{S}}_{w}^{t} = \frac{1}{G} \sum_{g=1}^G(\text{Concat}(\text{Attention}_1, \ldots, \text{Attention}_H) W_O) \label{eq:energy1}
\end{equation}
\end{minipage}
}
 
Our approach captures complex, non-linear relationships within and across series to enable holistic understanding, which is crucial for interpretation, decision-making, reliability in multi-sensor environments, and significantly enhancing forecast accuracy. We discovered spatial dependencies that go beyond pairwise relationships from the data. In the following section, we incorporate an accurate and reliable predefined graph, constructed using domain expert knowledge, to improve predictions.

\vspace{-3mm}
\subsubsection{Graph Chebyshev convolution} 
\vspace{-2mm} 
Graph convolution is an effective method for processing graph-structured data, with spectral graph convolution \cite{tanaka2021graph} being notable but computationally intensive. To address this, Chebyshev Graph Convolution (CGC) \cite{Defferrard2016} offers a more scalable alternative, leveraging Chebyshev polynomials to approximate spectral graph convolution, facilitating efficient convolutional filtering on graph-structured data using the Chebyshev polynomial approximation of the graph Laplacian. The Chebyshev polynomials are calculated based on the normalized Laplacian matrix of the predefined graph, denoted as $\hat{L} = \hat{D}^{-1/2} \hat{A} \hat{D}^{-1/2}$, where $\hat{A}$ is the normalized adjacency matrix, and $\hat{D}$ is the diagonal degree matrix of the graph. The Chebyshev approximation of the graph Laplacian to any degree is obtained using Chebyshev polynomials $T_k(\hat{L})$, where $k$ represents the degree of the polynomial.   The GCC operation can be defined as follows:

\vspace{0mm}
\resizebox{0.835\linewidth}{!}{
\begin{minipage}{\linewidth}
\begin{equation}
\mathbf{\dot{S}}_{w}^{t} = \sigma\left(\sum_{k=0}^{K-1} T_k(\hat{L}) \mathbf{\tilde{S}}_{w}^{t} \Theta_k\right) 
\label{eq:energy2}
\end{equation}
\end{minipage}
}

\vspace{0mm}
where $\sigma(\cdot)$ is a non-linear activation function applied element-wise, $\Theta_k \in \mathbb{R}^{d \times d}$ is the trainable weight matrix for the $k$-th order Chebyshev polynomial, and $K$ denotes the maximum order of the Chebyshev polynomials, which influences the expressive power of the approximation. $\mathbf{\dot{S}}_{w}^{t}$ is the transformed time series embedding, which captures the spatial relationships within the graph. To regulate the information flow from $\mathbf{\bar{S}}_{w}^{t}$ (refer Equation \eqref{eq:energy1}) and $\mathbf{\dot{S}}_{w}^{t}$ (refer Equation \eqref{eq:energy2}), we employ a gating mechanism that generates a weighted combination of these representations, denoted as $\mathbf{\widetilde{S}}_{w}^{t}$. Our hybrid architecture combines explicit domain-expert knowledge with implicit knowledge, extending beyond pairwise dependencies to capture the full complexity of spatio-temporal dependencies. Our approach, by modeling both local and global relationships in spatio-temporal data, enables accurate forecasting

\vspace{-3mm}
\subsubsection{Output Layer}
\vspace{-2mm}
We employ the multi-head attention mechanism (MHA) \cite{vaswani2017attention} to merge text-level and time series embeddings, thereby enhancing contextual understanding and alignment across different multi-domain embeddings. This integration improves the analysis and understanding of multi-time-series data, leading to more accurate framework predictions, $\hat{\mathbf{S}}^{t+1}$, by combining insights from both modalities.

\vspace{-3mm}
\subsubsection{Uncertainity Estimation}
\vspace{-1mm} 
We present a variant and extension of our proposed framework, \texttt{MultiTs} Net for time series forecasting: \texttt{w/Unc-MultiTs} Net, with a focus on uncertainty estimation. The \texttt{MultiTs} utilizes a supervised learning approach to minimize the Mean Absolute Error (MAE) which quantifies the deviation between the framework's forecasts and actual data. The \texttt{w/Unc-MultiTs} Net extends this by assessing uncertainties in forecasts, with predictions modeled as a heteroscedastic Gaussian distribution, characterized by mean $\mu_\phi\big(\mathbf{\widetilde{S}}_{w}^{t}\big)$ and variance $\sigma_\phi^2\big(\mathbf{\widetilde{S}}_{w}^{t}\big)$. These parameters are derived using $ \mu_\phi\big(\mathbf{\widetilde{S}}_{w}^{t}\big), \sigma_\phi^2\big(\mathbf{\widetilde{S}}_{w}^{t}\big) = f_\theta\big(\mathbf{\widetilde{S}}_{w}^{t}\big)$, from a linear layer function $f_\theta$ applied to the output of a multi-modal alignment output layer. 

\vspace{-2mm}
\resizebox{0.78\linewidth}{!}{
\begin{minipage}{\linewidth}
\begin{equation}
\mathcal{L}_{\text{GaussianNLLLoss}} = \sum_{t=1}^{\text{T}} \left[\frac{\log \sigma_\phi\big(\mathbf{\widetilde{S}}_{w}^{t}\big)^2}{2}+\frac{\left(\mathbf{S}^{t+1}-\mu_\phi\left(\mathbf{\widetilde{S}}_{w}^{t}\right)\right)^2}{2 \sigma_\phi\left(\mathbf{\widetilde{S}}_{w}^{t}\right)^2}\right] \nonumber
\end{equation}
\end{minipage}
}

The framework minimizes the Gaussian negative log-likelihood loss, thereby enhancing the quantification of uncertainty. While the \texttt{MultiTs} Net focuses on minimizing MAE for accurate forecasting, the \texttt{w/Unc-MultiTs} Net emphasizes minimizing the Gaussian negative log-likelihood loss for effective uncertainty quantification in time series forecasting.

\vspace{-3mm}
\subsubsection{Irregular Time Series}
\vspace{-2mm}
We focus on evaluating the effectiveness of the \texttt{MultiTs} Net framework in handling missing data in large, complex sensor networks. The study simulates two types of missingness patterns: MCAR (Missing Completely At Random), representing random sensor failures, and block-missing, where data points are missing for a contiguous period. Block-missing patterns are defined by their length (the number of missing points) and frequency (how often they occur). These simulations help assess the framework's performance in 12-sequence-to-24-sequence forecasting tasks, with data missingness ranging from 10$\%$ to 50$\%$. Tables \ref{tab:pfe1} and \ref{tab:pfe2} show the imputation results. The performance of the proposed models in the forecasting task was assessed by calculating the average error between predicted and actual values over 12 future time steps. The results reveal that while the \texttt{MultiTs} Net performs well with lower missing data percentages, its accuracy declines as the level of missingness increases. However, its ability to condition forecasts on available observations without relying on imputed values demonstrates its resilience and effectiveness in capturing nonlinear spatio-temporal dependencies in sensor network data.

\vspace{-2mm}
\begin{table*}[ht!]
\setlength{\tabcolsep}{0.275em} 
\renewcommand\arraystretch{1.275} 
\centering
\resizebox{1.0\textwidth}{!}{
\hspace*{-0.5cm}\begin{tabular}{c|c|ccc|ccc|ccc|ccc}
\hline
\multirow{2}{*}{\textbf{Missing Scheme}} & \multirow{2}{*}{\textbf{Missing Rate}} & \multicolumn{3}{c|}{\textbf{PeMSD3}} & \multicolumn{3}{c|}{\textbf{PeMSD4}} & \multicolumn{3}{c|}{\textbf{PeMSD7}} & \multicolumn{3}{c}{\textbf{METR-LA}} \\ \cline{3-14} 
 &  & \textbf{RMSE} & \textbf{MAE} & \textbf{MAPE} & \textbf{RMSE} & \textbf{MAE} & \textbf{MAPE} & \textbf{RMSE} & \textbf{MAE} & \textbf{MAPE} & \textbf{RMSE} & \textbf{MAE} & \textbf{MAPE} \\ \hline
\textbf{MultiTs Net} & \textbf{0\%} & 21.03 & 13.64 & 11.30 & 27.04 & 18.02 & 10.31 & 30.05 & 19.52 & 8.08 & 7.53 & 4.56 & 9.47 \\ \hline
\multirow{3}{*}{Point} & 10\% & 21.83 & 13.50 & 11.97 & 26.97 & 18.11 & 12.01 & 31.24 & 21.79 & 8.55 & 7.92 & 4.87 & 8.63 \\
 & 30\% & 21.99 & 13.48 & 12.83 & 31.07 & 20.05 & 11.95 & 30.66 & 22.05 & 9.09 & 8.47 & 5.46 & 9.64 \\
 & 50\% & 22.08 & 15.46 & 12.67 & 32.10 & 20.65 & 13.70 & 32.41 & 21.91 & 10.35 & 8.77 & 5.82 & 9.29 \\ \hline
 \multirow{3}{*}{Block} & 10\% & 19.87 & 13.70 & 11.82 & 28.27 & 18.39 & 11.52 & 29.84 & 20.65 & 8.39 & 8.40 & 4.52 & 8.86 \\
 & 30\% & 21.14 & 13.96 & 12.65 & 29.59 & 20.34 & 12.43 & 30.85 & 22.33 & 8.78 & 8.73 & 5.05 & 9.57 \\
 & 50\% & 23.01 & 14.72 & 13.88 & 33.18 & 20.88 & 13.81 & 33.04 & 23.21 & 10.21 & 9.47 & 5.56 & 9.39 \\ \hline
\end{tabular}
}
\vspace{-3mm}
\caption{The table shows the pointwise forecasting error on 12-sequence-24-sequence forecasting task on irregular datasets.}
\label{tab:pfe1}
\vspace{-2.5mm}
\end{table*}

\vspace{-2mm}
\begin{table*}[ht!]
\setlength{\tabcolsep}{0.425em} 
\renewcommand\arraystretch{1.20} 
\centering
 \resizebox{0.9\textwidth}{!}{
\begin{tabular}{c|c|ccc|ccc|ccc}
\hline
\multirow{2}{*}{\textbf{Missing Scheme}} & \multirow{2}{*}{\textbf{Missing Rate}} & \multicolumn{3}{c|}{\textbf{PeMSD7(M)}} & \multicolumn{3}{c|}{\textbf{PeMSD8}} & \multicolumn{3}{c}{\textbf{PEMS-BAY}} \\ \cline{3-11} 
 &  & \textbf{MAE} & \textbf{RMSE} & \textbf{MAPE} & \textbf{MAE} & \textbf{RMSE} & \textbf{MAPE} & \textbf{MAE} & \textbf{RMSE} & \textbf{MAPE} \\ \hline
\textbf{MultiTs Net} & \textbf{0\%} & 4.97 & 2.71 & 5.57 & 21.49 & 13.85 & 7.80 & 2.81 & 1.50 & 2.73 \\ \hline
\multirow{3}{*}{Point} & 10\% & 5.16 & 2.97 & 6.85 & 22.61 & 15.91 & 8.67 & 2.82 & 1.60 & 3.20 \\
 & 30\% & 5.60 & 3.63 & 7.63 & 24.67 & 16.96 & 9.48 & 3.04 & 1.77 & 3.45 \\
 & 50\% & 5.93 & 3.69 & 7.90 & 26.96 & 17.50 & 9.90 & 3.16 & 2.01 & 3.44 \\ \hline
\multirow{3}{*}{Block} & 10\% & 5.20 & 3.15 & 6.88 & 23.03 & 15.59 & 9.13 & 2.98 & 1.68 & 3.10 \\
 & 30\% & 5.53 & 3.49 & 7.65 & 24.16 & 16.48 & 9.80 & 3.11 & 1.79 & 3.31 \\
 & 50\% & 5.82 & 3.76 & 8.18 & 26.12 & 17.96 & 10.70 & 3.25 & 1.92 & 3.50 \\ \hline 
\end{tabular}
}\\[-1.5ex]
\caption{The table shows the pointwise forecasting error on 12-sequence-24-sequence forecasting task on irregular datasets.}
\label{tab:pfe2}
\end{table*}

\newpage
\vspace{-5mm}
\newpage
\vspace{-1mm}

\pagebreak
\newpage
\bibliography{iclr2024_conference}
\bibliographystyle{iclr2024_conference}

\end{document}